\documentclass[sigplan,10pt]{acmart}
\renewcommand\footnotetextcopyrightpermission[1]{}

\usepackage{tikz}
\usepackage{amsmath}
\usepackage[normalem]{ulem}
\usepackage{endnotes,microtype,xspace,fancyvrb,multirow}
\usepackage{xcolor}
\usepackage{xspace}
\usepackage{balance}
\usepackage{pifont}
\usepackage{anyfontsize}
\usepackage{booktabs}
\usepackage{fp}
\usepackage{caption}
\usepackage[frozencache,cachedir=minted-cache]{minted}
\usepackage{listings}
\usepackage{soul}
\usepackage{balance}
\usepackage{textcomp}
\usepackage{stmaryrd}
\usepackage{enumitem}
\usepackage{array}
\usepackage{wrapfig}
\usepackage{makecell}
\usepackage{subcaption}
\usepackage{tabularx}
\usepackage{fontenc}
\usepackage{bbding}  
\usepackage{libertine}
\usepackage{xcolor}
\usepackage{colortbl}

\newif\ifArxiv
\Arxivtrue

\usepackage{xparse}

\newcommand{\histopipe}{Histo\-Pipe\xspace}
\newcommand{\histospec}{Histo\-Spec\xspace}

\ifArxiv
    \newcommand{\sys}{Rhyme\-RL\xspace}
    
    \newcommand{\gpuSHORT}{}
    \newcommand{\gpu}{high-performance\xspace}
    \newcommand{\mytitle}{History Rhymes: Accelerating LLM Reinforcement Learning with \sys}
    \newcommand{\myauthor}{\rm Jingkai He$^{1}$,\; Tianjian Li$^{2}$,\; Erhu Feng\textsuperscript{\Envelope}$^{1}$,\; Dong Du\textsuperscript{\Envelope}$^{1}$,\; Qian Liu$^{2}$,\; Tao Liu$^{2}$,\; \\Yubin Xia$^{1}$,\; Haibo Chen$^{1}$\\
    {\normalsize {$^1$Institute of Parallel and Distributed Systems, Shanghai Jiao Tong University}} \\
    {\normalsize {$^2$ByteDance}}}
    \newcommand{\noArxiv}[1]{}
\fi

\definecolor{hjkcolor}{RGB}{85,107,47}

\newcommand{\codeword}[1]{\textcolor{black}{$\mathsf{#1}$}}

\newcommand{\Ddhighlight}[1]{\textbf{\color{hightcode}{$\blacktriangleright$#1}}}
\newcommand{\Ddhighlightcomment}[1]{\textbf{\color{hightcomment}{$\blacktriangleright$#1}}}
\newcommand{\Ddhighlightcommentnotri}[1]{\textbf{\color{hightcomment}{#1}}}
\newcommand{\Ddhighlightnotri}[1]{\textbf{\color{hightcode}{#1}}}
\newcommand{\highlightfuncword}[1]{{\color[HTML]{850a57}{{#1}}}}

\newcommand{\Ddhighlightfuncname}[1]{{\textbf{\color[HTML]{000080}{#1}}}}
\newcommand{\Ddhighlightfunc}[1]{\textbf{\color{diaozi}{{#1}}}}

\newcommand{\myparagraph}[1]{\noindent\textbf{#1}}

\DeclareCaptionFont{myfont}{\fontsize{9.4pt}{11.28pt}\selectfont}
\DeclareCaptionFont{subfigfont}{\fontsize{9pt}{10.8pt}\selectfont}
\definecolor{diaozi}{RGB}{93,49,49}
\definecolor{hightcode}{rgb}{1.0,0.13,0.32}
\definecolor{hightcomment}{RGB}{205, 92, 92}
\definecolor{myRubineRed}{RGB}{209, 0, 86}

\newcommand{\cmark}{\color[HTML]{32CB00} \ding{51}}%
\newcommand{\xmark}{\color[HTML]{FE0000} \ding{55}}%

\newcommand{\showfontsize}{The current font size is: \the\fontdimen6\font}
\newcommand{\showlinespread}{The current line spread is: \the\baselineskip}

\begin{document}

\date{}

\title{\mytitle}

\author{\myauthor}

\settopmatter{printacmref=false}
\settopmatter{printfolios=true}
\setcopyright{none}

\begin{abstract}

With the rapid advancement of large language models (LLMs), reinforcement learning (RL) 
has emerged as a pivotal methodology for enhancing the reasoning capabilities of LLMs. 
Unlike traditional pre-training approaches, RL encompasses multiple stages: \emph{rollout}, \emph{reward}, and \emph{training}, 
which necessitates collaboration among various worker types.
However, current RL systems continue to grapple with substantial GPU underutilization, due to two primary factors:
(1) The rollout stage dominates the overall RL process due to test-time scaling;
(2) Imbalances in rollout lengths (within the same batch) result in GPU bubbles.
While prior solutions like asynchronous execution and truncation offer partial relief, they may compromise training accuracy for efficiency.

Our key insight stems from a previously overlooked observation: \emph{rollout responses exhibit remarkable similarity across adjacent training epochs}.
Based on the insight, we introduce \sys, an LLM RL system designed to accelerate RL training 
with two key innovations.
First, to enhance rollout generation, we present \histospec, a speculative decoding inference engine that utilizes the similarity of historical rollout token sequences to obtain accurate drafts. 
Second, to tackle rollout bubbles, we introduce \histopipe, a two-tier scheduling strategy that leverages the similarity of historical rollout distributions to balance workload among rollout workers.
We have evaluated \sys within a real production environment, 
demonstrating scalability from dozens to thousands of GPUs. 
Experimental results demonstrate that \sys achieves a 2.6x performance improvement over existing methods, 
without compromising accuracy or modifying the RL paradigm.

\end{abstract}

\maketitle
\pagestyle{plain}

\section{Introduction}
\label{s:intro}

Post-training with reinforcement learning (RL) has emerged as a new paradigm for scaling and enhancing the capabilities of LLMs~\cite{deepseekai2025deepseekr1incentivizingreasoningcapability,comanici2025gemini25pushingfrontier,llama4,yang2025qwen3technicalreport,claude4}. 
Representative post-training models, such as DeepSeek-R1~\cite{deepseekai2025deepseekr1incentivizingreasoningcapability}, have demonstrated significant improvements in areas including coding~\cite{claude4}, mathematics~\cite{shao2024deepseekmathpushinglimitsmathematical} and many others~\cite{zheng2025deepresearcherscalingdeepresearch,wu2025agenticreasoningstreamlinedframework,prabhakar2025omnisciencedomainspecializedllmscientific,zhou2025sweetrltrainingmultiturnllm}.
A standard RL pipeline comprises three main stages: \emph{rollout}, \emph{reward}, and \emph{training}. 
In the rollout stage, the LLM generates a large number of tokens,
with extended reasoning to improve the quality of final responses (test-time scaling~\cite{muennighoff2025s1simpletesttimescaling}). 
In the reward stage, a reward score is assigned to each response (i.e., sample).
In the subsequent training stage, the model's weights are updated by computing new loss values based on the assigned rewards.
As a result, RL systems are inherently complex and distributed, 
typically involving coordination among multiple heterogeneous workers to efficiently execute the various pipeline stages.

Current LLM RL systems face significant GPU underutilization,
primarily arising from two sources.
First, \emph{overly long rollout phases.} During each RL step, the rollout phase, which entails generating a substantial number of thinking tokens,
typically consumes 84\% to 91\% of the total time. 
Additionally, the autoregressive nature of LLMs prevents the rollout phase from fully utilizing GPU computational resources, 
resulting in substantial underutilization.
Second, \emph{bubbles caused by imbalanced rollouts.} Within a single batch, the response lengths of rollouts generated by different prompts vary dramatically, 
leading to a pronounced long-tail effect.
As a result, completed rollout tasks must wait until the longest rollout in the batch finishes before advancing to reward and training stages.
Due to these factors, we observe that SOTA RL systems, such as veRL\,\cite{10.1145/3689031.3696075}, experience over 46\% GPU resource idleness, 
significantly compromising the efficiency of RL.

To address these problems, recent research has mainly focused on scheduling optimizations.
Some systems mitigate the performance degradation caused by long-tail rollouts by truncation (e.g., Kimi-K2~\cite{kimiteam2025kimik2openagentic}) or reducing their batch size (e.g., StreamRL~\cite{zhong2025streamrlscalableheterogeneouselastic}).
However, truncation methods inherently introduce a trade-off between accuracy and efficiency, 
while adjusting batch sizes provides only limited alleviation of the long-tail problem ($\sim$10\%~\cite{zhong2025streamrlscalableheterogeneouselastic}). 
Some systems (e.g., AReaL~\cite{fu2025areallargescaleasynchronousreinforcement}) depart from conventional RL systems by enabling full asynchronization between the rollout and training phases to reduce GPU idle time. However, this results in rollouts utilizing stale model weights, which changes the paradigm of RL. %
Additionally, model weight updates during rollout trigger the recomputation of all tokens in the ongoing rollouts, leading to considerable overhead.
Therefore, enhancing the efficiency of rollout generation and minimizing GPU resource idleness remain critical challenges in the development of LLM RL systems.

Through a detailed analysis of the real-world RL training,
we observe that the rollout process in RL is distinct from conventional LLM inference.
Specifically, a complete RL training consists of performing multiple inference passes on the same prompt in different RL steps (50---100 \emph{epochs}~\cite{shao2024deepseekmathpushinglimitsmathematical,yu2025dapoopensourcellmreinforcement,deepscaler2025}). 
While the model weights vary between these steps due to continuous updates during training, 
current RL algorithms (such as GRPO~\cite{shao2024deepseekmathpushinglimitsmathematical}, etc.~\cite{yu2025dapoopensourcellmreinforcement, zheng2025groupsequencepolicyoptimization}) apply \emph{clipping operations}~\cite{huang2024ppoclipattainsglobaloptimality}, which restrict the magnitude of the model's updates at each step, to maintain stable model evolution. 
This stability leads to a \emph{high degree of similarity} between rollout responses across different epochs:
(1) \emph{Token sequence similarity.}
The responses generated for each prompt exhibit substantial similarity to their corresponding previous rollout responses, with 75\%---95\% of tokens being reusable.
(2) \emph{Length distribution similarity.} Although a prompt generates responses of varying lengths in different epochs, their position within the epoch's response length ranking remains stable, with only 2\%-4\% of responses experiencing significant rank changes. 
Motivated by this observation, our central insight is to \emph{exploit the \textbf{similarity from historical rollouts} to accelerate the rollout process 
and achieve effective load balancing across rollout workers.}

\begin{figure}[t]
    \centering
    \includegraphics[width=\linewidth]{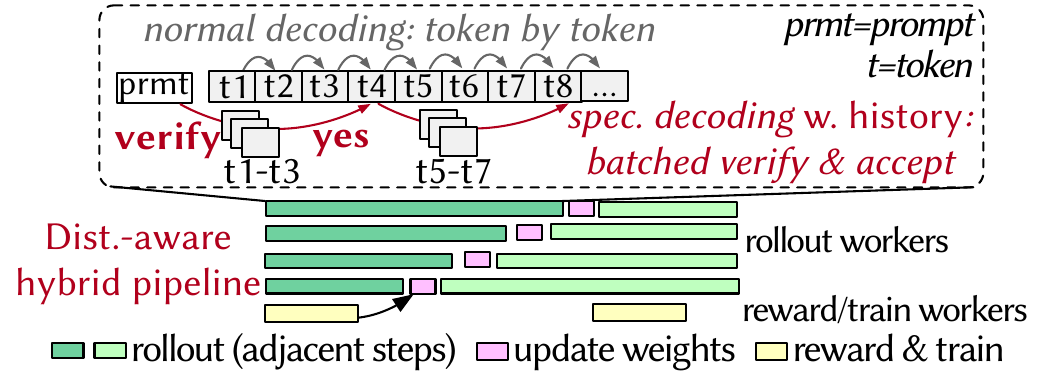}
    \setlength{\belowcaptionskip}{-10pt}
	\setlength{\abovecaptionskip}{-5pt}
    \caption{\textbf{An overview of \sys's designs.}
    }
    \label{fig:intro-overview}
\end{figure}

We propose a novel RL system, \sys, 
which significantly improves LLM RL efficiency without modifying existing RL training paradigms or sacrificing accuracy, as \autoref{fig:intro-overview} shows.
First, leveraging the historical token sequence similarity, we propose a lightweight yet effective \emph{speculative decoding}~\cite{10.5555/3618408.3619203} mechanism, \emph{\histospec}, to accelerate the rollout process and improve the computational density.
\histospec adopts a reward-aware, tree-based management strategy to organize draft tokens,
and incorporates a novel speculative strategy inspired by TCP congestion control mechanisms~\cite{896303} to improve prediction accuracy.
Secondly, leveraging the historical length distribution similarity,
we introduce the \emph{\histopipe} scheduling.
By complementing long and short rollouts between adjacent steps, 
\histopipe effectively balances workload across rollout workers and eliminates GPU bubbles caused by the imbalanced distribution of rollout lengths.
\histopipe further tackles outliers with migration-based rebalancing, and proposes a two-tier scheduling mechanism for better complementarity.

We have implemented our system on veRL and successfully deployed it in industrial RL environments ranging from dozens to over a thousand GPUs, 
achieving a performance improvement of up to 2.6x. \sys consistently attains SOTA performance across GPU clusters of varying scales
while maintaining training accuracy without compromise.

\sys will be open source.

\section{Background}
\label{s:back}

\subsection{LLM Reinforcement Learning}
RL-based post-training refines LLMs through interaction with environments. 
As the marginal benefits of pretraining diminish, RL is widely acknowledged as a pivotal approach for enhancing LLMs' reasoning capabilities~\cite{deepseekai2025deepseekr1incentivizingreasoningcapability,comanici2025gemini25pushingfrontier, openai2024openaio1card, gpto3}.

\begin{figure}[t]
    \centering
    \includegraphics[width=0.7\linewidth]{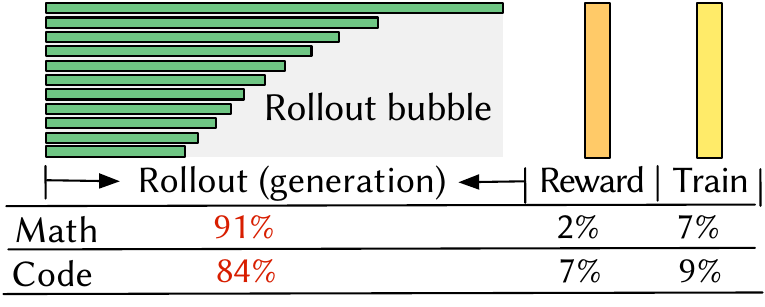}
    \setlength{\belowcaptionskip}{-5pt}
	\setlength{\abovecaptionskip}{0pt}
    \caption{\textbf{Phases and time distribution in LLM RL.}
    \emph{We train 32B models with math/code datasets using veRL~\cite{10.1145/3689031.3696075} and GRPO~\cite{shao2024deepseekmathpushinglimitsmathematical}.}
    }
    \label{fig:motiv-rl-stages}
\end{figure}

\myparagraph{Workflow.} A whole RL process consists of multiple repeated steps. In brief, each step comprises three stages, as \autoref{fig:motiv-rl-stages} shows:
(1) \emph{\textit{Rollout}}: the LLM generates responses to input prompts in batches (i.e., LLM inference).
(2) \emph{\textit{Reward}}: The responses are scored (i.e., rewarded) using rule-based functions (e.g., running tests for code) or reward models.
(3) \emph{\textit{Train}}: Loss is computed based on rewards, followed by backpropagation to optimize the model and generate new model weights. 

\myparagraph{Algorithm.} Early algorithms like PPO~\cite{schulman2017proximalpolicyoptimizationalgorithms,pmlr-v37-schulman15} employ not only rule-based functions/reward models to generate sample-level rewards, but also simultaneously train a critic model to produce action-level rewards. This strategy stabilized training by reducing the volatility of rewards from sample-level evaluations. Contemporary mainstream methods, exemplified by GRPO~\cite{shao2024deepseekmathpushinglimitsmathematical} and DAPO~\cite{yu2025dapoopensourcellmreinforcement}, have superseded critic models with \emph{Group Relative Advantage (GRA)}. This paradigm leverages the inherent stochasticity of LLMs during rollout: for each prompt, the LLM generates \emph{a group of responses} (usually 16 in our production). Policy updates are then guided by relative scoring within the groups. The process of generating multiple responses per prompt can be interpreted as the model exploring diverse solution pathways to a problem.

\subsection{Speculative Decoding}
\label{sec:spec-back}

During LLM decoding, each attention operation requires accessing the entire KV cache, making the decoding process memory bandwidth-bound and preventing full utilization of computational resources. Unlike conventional decoding, which generates one token per iteration (i.e., LLM forward pass), speculative decoding~\cite{10.5555/3618408.3619203, 10.1145/3620666.3651335, zhang2024accelerating, chen2023acceleratinglargelanguagemodel, liao2025rewardguidedspeculativedecodingefficient, du2025reinforcementspeculativedecodingfast} first predicts multiple draft tokens using a lightweight method (e.g., using a small draft model~\cite{cai2024medusasimplellminference,li2025eaglespeculativesamplingrequires,li2024eagle2fasterinferencelanguage, li2025eagle3scalinginferenceacceleration} or retrieving from corpora~\cite{oliaro2025suffixdecodingextremespeculativedecoding,saxena2023prompt, yang2023inferencereferencelosslessacceleration}). 
The LLM then verifies these tokens in a single forward pass by computing their logits and accepting valid tokens. 
Since verifying multiple tokens incurs nearly identical memory access overhead (traversing the KV cache once) as generating one token conventionally, but with higher computational intensity, speculative decoding amortizes the memory cost across multiple tokens. This approach is theoretically proven to preserve output distribution integrity. When acceptance rates are favorable, it significantly accelerates LLM decoding~\cite{How-Speculative-Decoding-Boosts-vLLM}.

\sys is the first LLM RL system leveraging speculative decoding to accelerate the time-consuming rollout process.

\section{Characterising RL Training in the Wild}
\label{s:overview}

\begin{figure*}[t]
    \centering
    \begin{subfigure}[b]{0.33\textwidth}
        \centering
        \includegraphics[width=\textwidth]{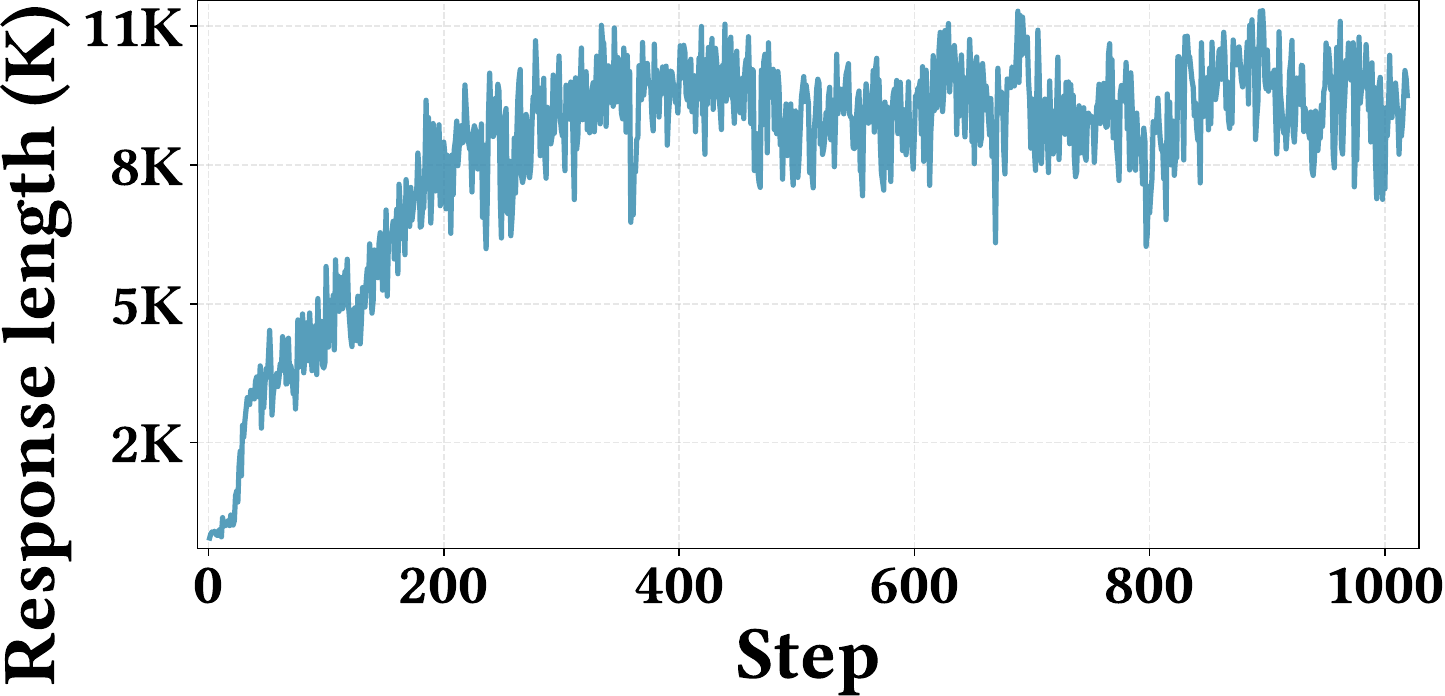}
        \caption{\textbf{Mean response lengths across steps}}
        \label{fig:rollout-dominant}
    \end{subfigure}
    \hfill
    \begin{subfigure}[b]{0.33\textwidth}
        \centering
        \includegraphics[width=\textwidth]{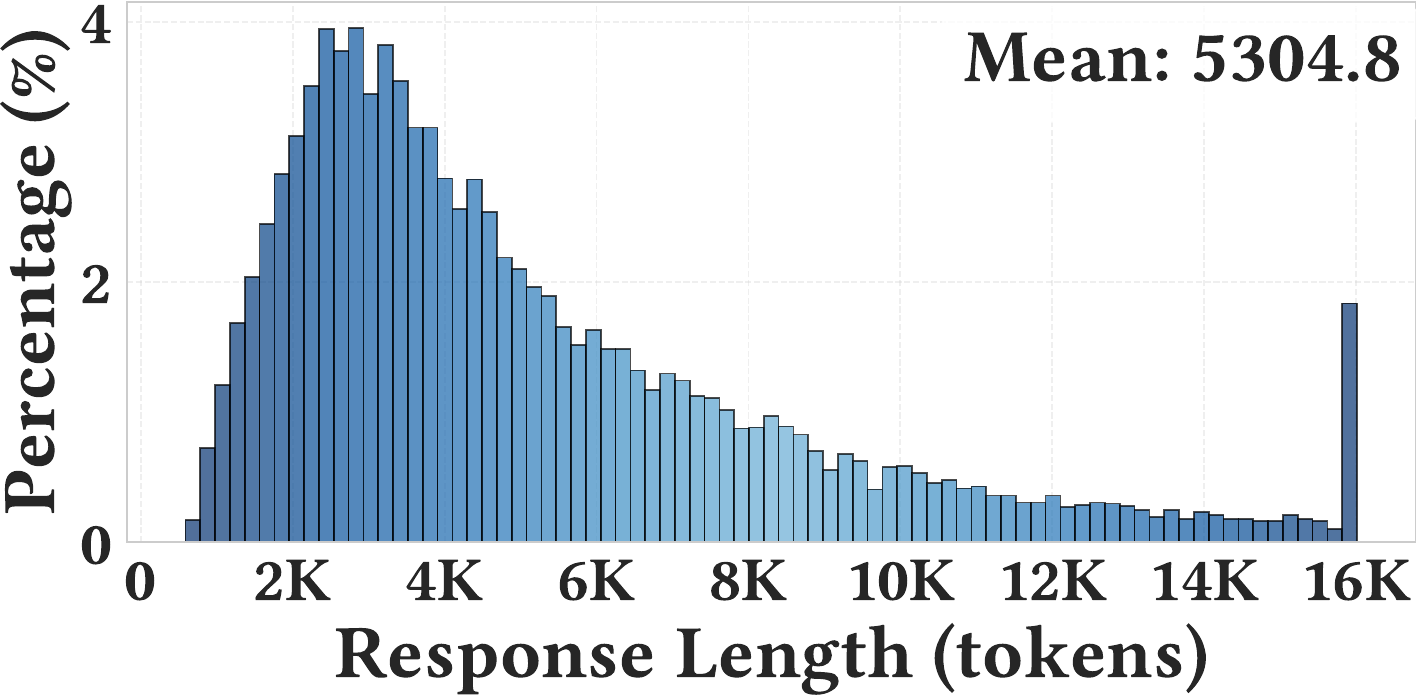}
        \caption{\textbf{Response length distribution in a step}}
        \label{fig:response-len-distri}
    \end{subfigure}
    \begin{subfigure}[b]{0.33\textwidth}
        \centering
        \includegraphics[width=\textwidth]{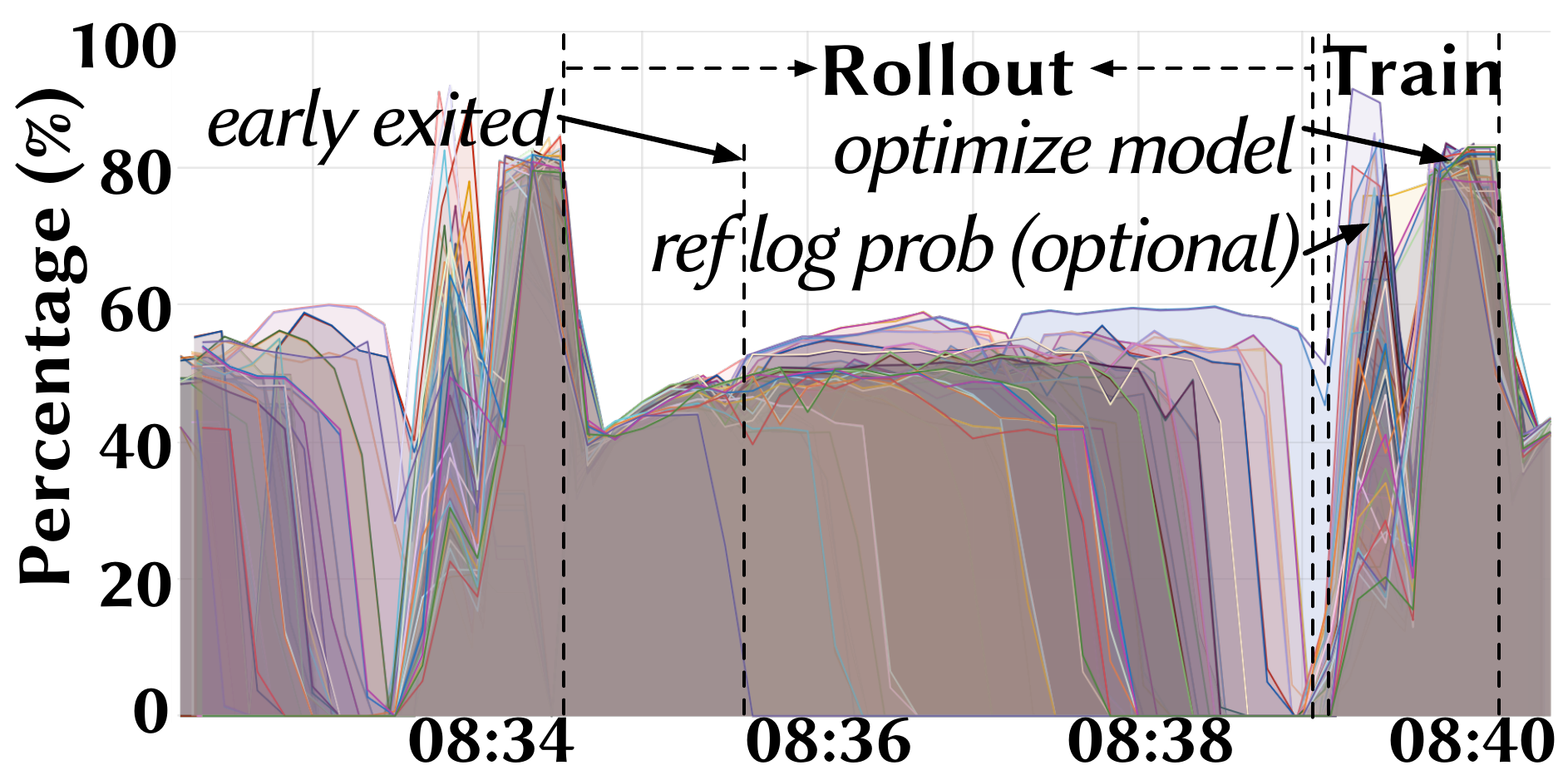}
        \caption{\textbf{Real-world GPU SM utilization}}
        \label{fig:intro-gpu-utilize}
    \end{subfigure}
    \setlength{\belowcaptionskip}{-5pt}
	\setlength{\abovecaptionskip}{-10pt}
    \caption{\textbf{Diving into real-world RL training of a 32B LLM.}
        \emph{
    We train the model using math datasets (>200K samples) with 64 GPUs. The max response length is set to 16K tokens.
    In Fig.-c, each line represents a GPU's SM utilization. 
    The training stage can be further divided into ref log prob computing (optional), which computes the generated tokens' logits via a forward pass of the reference LLM, and model optimizing.
    }
 }
    \label{fig:rl-workload-characteristics}
\end{figure*}

Despite the widespread adoption of RL for LLM training,
current RL systems still face significant performance challenges.
This section presents the key implications derived from our practical experience in training state-of-the-art LLMs and analysis of real-world traces.

\subsection{Rollout as the Major Bottleneck in RL Training}

\myparagraph{Implication-1: Rollout dominates the RL training timeline.}
During RL training, we observe that the rollout phase dominates the RL time.
Specifically, as \autoref{fig:motiv-rl-stages} shows,
for LLMs trained with a maximum response length of 16K tokens, rollout accounted for 91\% of the entire RL processing time for the math model and 84\% for the code model. When the maximum response length was increased to 32K tokens or longer, the rollout time overhead exceeded 95\%.

This significant overhead stems from three key factors:
(1) \textit{Complex reasoning demands.} During RL, LLMs are required to generate complex reasoning chains in responses. This results in long responses, and crucially, the response length tends to increase progressively as training advances.
As \autoref{fig:rollout-dominant} shows, the mean response length grows from 1K to 10K in 320 steps.
(2) \textit{Memory-bound decoding.} During rollout, generating each token requires an LLM forward pass, which is constrained by memory bandwidth~\cite{10.1145/3575693.3575747}.
(3) \textit{Sequential dependency.} Due to the dependency between rollout and the subsequent training step, the training phase cannot commence until the longest sequence within the batch completes its rollout.

\myparagraph{Implication-2: GPU underutilization from rollout imbalance.}
Preserving the rollout-train dependency is critical for training accuracy, but it introduces significant compute bubbles and
leads to GPU idleness during rollout.
We observe a significant long-tail effect 
within a batch, as shown in \autoref{fig:response-len-distri} --- rollout response lengths vary widely across sequences. 
Due to the imbalanced distribution of rollout lengths, some rollout workers (i.e., GPUs cooperating with tensor parallelism) finish early and become idle, yet must remain inactive until all workers complete their tasks.
As \autoref{fig:intro-gpu-utilize} shows, in a step, GPU monitoring reveals that the earliest-finishing GPU remains idle for $\sim$76\% of the total rollout duration.

\subsection{State-of-the-Art Efforts and the Limitations}
\label{sec:existing-work}
We summarize existing efforts on optimizing the efficiency of LLM RL systems~\cite{10.1145/3689031.3696075, kimiteam2025kimik2openagentic, deepscaler2025, zhong2025streamrlscalableheterogeneouselastic, fu2025areallargescaleasynchronousreinforcement, mei2025real, han2025asyncflowasynchronousstreamingrl, wang2025distflowfullydistributedrl, 305943, nemo, hu2025reinforceefficientrlhfalgorithm, wang2025reinforcementlearningoptimizationlargescale} in \autoref{tab:systems}.

\begin{table}[]
	\centering
    \setlength\tabcolsep{0.2\tabcolsep}
    \fontsize{8pt}{9.6pt}\selectfont
	\begin{tabular}{l|ccc|l}
		\cline{1-5}
		\textbf{Systems} &  \textbf{\begin{tabular}[c]{@{}c@{}}Reduce \\ rollout \\ time\end{tabular}}  &  \textbf{\begin{tabular}[c]{@{}c@{}}Tackle \\ rollout \\bubbles\end{tabular}}  & \textbf{\begin{tabular}[c]{@{}c@{}}Rollout\\-train \\ pipeline\end{tabular}} & \textbf{Core techniques}\\
		\cline{1-5}
		HybridFlow\,\cite{10.1145/3689031.3696075}& \xmark& \xmark& \xmark & Hybrid programming model\\
		Kimi K2\,\cite{kimiteam2025kimik2openagentic}& \xmark& Partially& \xmark & Partial rollout\\
		DeepScaleR\,\cite{deepscaler2025}& \xmark& \xmark& \cmark & \begin{tabular}[l]{@{}l@{}}Pipelined rollout/train\end{tabular}\\
		StreamRL\,\cite{zhong2025streamrlscalableheterogeneouselastic}& \xmark& Partially& \cmark & Skewness-aware scheduling\\
		AReaL\,\cite{fu2025areallargescaleasynchronousreinforcement,mei2025real}& \xmark& \cmark& \cmark & Fully async rollout\\
		AsyncFlow\,\cite{han2025asyncflowasynchronousstreamingrl}& \xmark& Partially& \cmark & Streaming pipeline\\
		DistFlow\,\cite{wang2025distflowfullydistributedrl}& \xmark& \xmark& \cmark & Distributed multi-controller\\
		\rowcolor{gray!20}\sys& \cmark& \cmark& \cmark & \begin{tabular}[c]{@{}c@{}}\histospec + \histopipe\end{tabular}\\
		\cline{1-5}
	\end{tabular} \\[0pt]
    \setlength{\belowcaptionskip}{0pt}
	\setlength{\abovecaptionskip}{-5pt}
	\caption{\textbf{Overview of existing systems optimizing LLM RL.}
	}
	\label{tab:systems}
\end{table}

\begin{figure}[t]
    \centering
    \begin{subfigure}[b]{\linewidth}
        \centering
        \includegraphics[width=\linewidth]{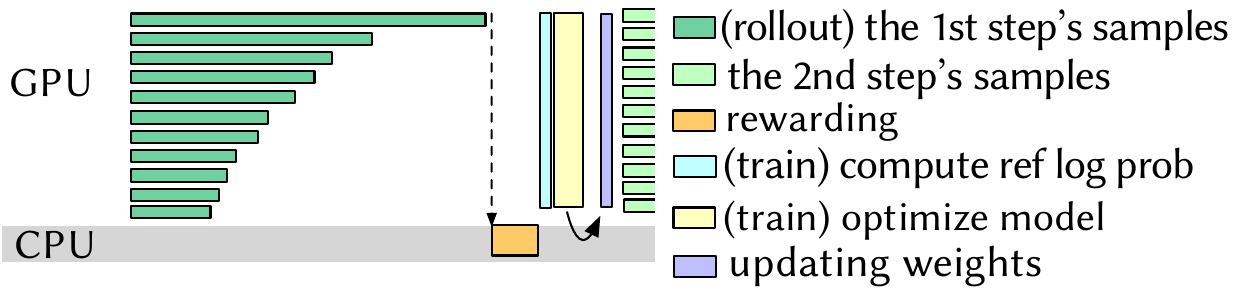}
        \caption{\textbf{Colocated RL training}}
        \label{fig:related-a}
    \end{subfigure}
    \hfill
    \begin{subfigure}[b]{\linewidth}
        \centering
        \includegraphics[width=\linewidth]{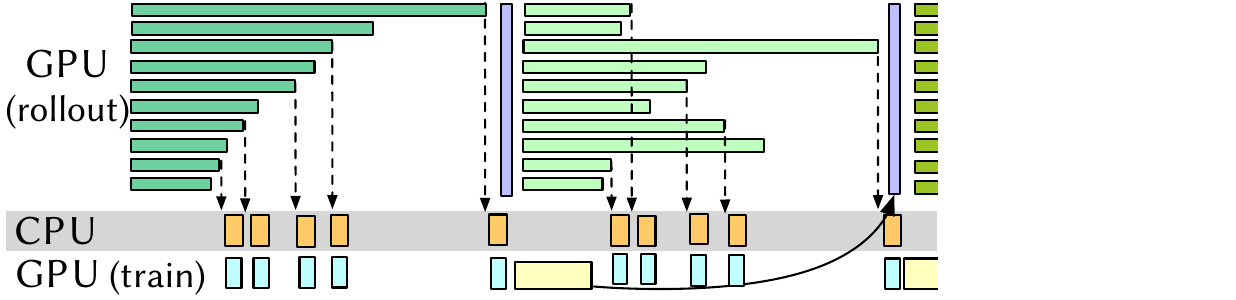}
        \caption{\textbf{Rollout-train streaming pipelined RL training}}
        \label{fig:related-b}
    \end{subfigure}
    \begin{subfigure}[b]{\linewidth}
        \centering
        \includegraphics[width=\linewidth]{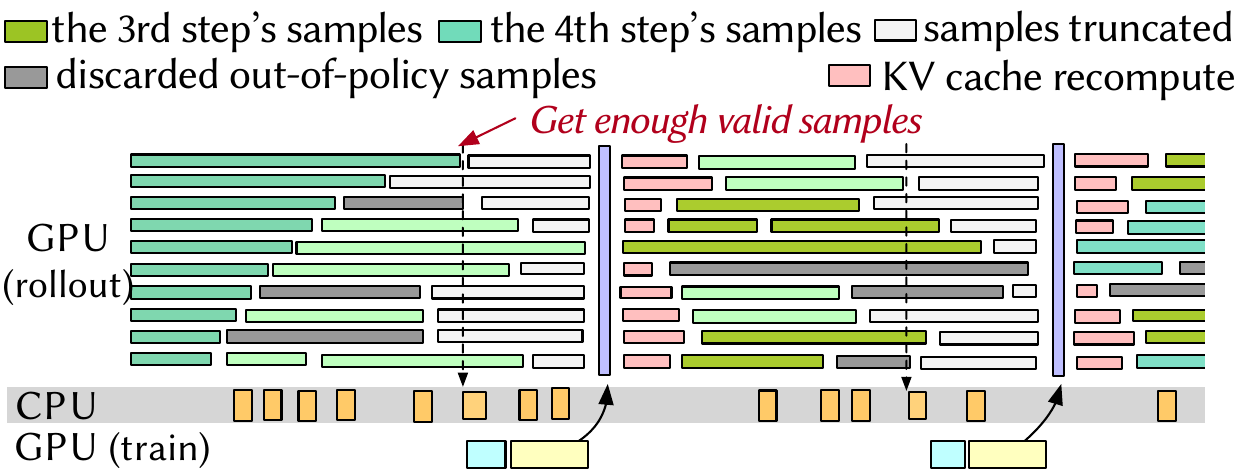}
        \caption{\textbf{Fully asynchronous rollout RL (max off-policyness\,=\,1)}}
        \label{fig:related-c}
    \end{subfigure}
    \setlength{\belowcaptionskip}{-15pt}
	\setlength{\abovecaptionskip}{-5pt}
    \caption{\textbf{Existing LLM RL pipelines.}
    \emph{
    Dashed lines represent the dependency between rollout samples and reward/train, while solid lines represent the dependency between training and weight updating. 
    }
 }
    \label{fig:related-pipeline}
\end{figure}

\myparagraph{Pipelining rollout and training.}
Early RL systems like veRL~\cite{10.1145/3689031.3696075} and OpenRLHF~\cite{hu2025reinforceefficientrlhfalgorithm} adopt a \emph{colocated architecture}, where GPUs repeatedly switch between rollout and training workloads, as \autoref{fig:related-a} shows. This design incurs significant overhead from context switching between workers and substantial bubbles due to strict step-wise rollout-train dependency.
Consequently, DeepScalaR~\cite{deepscaler2025} introduces a \emph{decoupled architecture}, relaxing the dependency by using stale weights that are one step behind for rollout (i.e., the off-policyness is 1\footnote{The off-policyness refers to the temporal discrepancy (in steps) between the model used to generate rollouts and the current model being optimized.}),
which is widely proven to maintain training accuracy~\cite{zhong2025streamrlscalableheterogeneouselastic,deepscaler2025,verl-one-step-off}.
This enables coarse-grained rollout-train pipeline, and is adopted by systems including veRL~\cite{10.1145/3689031.3696075,verl-one-step-off}.
AsyncFlow~\cite{han2025asyncflowasynchronousstreamingrl} further refines the rollout-train pipeline by granularizing dependencies from batch-level to mini-batch level, enhancing pipeline efficiency (as \autoref{fig:related-b} shows, the reward and train processes proceed after a mini-batch's rollout finishes). 
Nevertheless, imbalance persists among rollout workers, leaving significant bubbles within rollout stages.

\myparagraph{Tackling long-tail rollout}.
StreamRL~\cite{zhong2025streamrlscalableheterogeneouselastic} proposes \emph{skewness-aware scheduling}, which allocates additional data-parallel GPUs to the prompts likely to generate long responses. 
By reducing batch sizes for these prompts, it alleviates the long-tail effect in rollouts. However, due to the autoregressive nature of LLM rollouts, this approach only marginally reduces the long-tail and the rollout bubbles ($\sim$10\%~\cite{zhong2025streamrlscalableheterogeneouselastic}).
Kimi-K2~\cite{kimiteam2025kimik2openagentic} introduces \emph{partial rollout}, which truncates excessively long responses and retains generated segments for continuation in subsequent steps. 
This method faces an efficiency-accuracy dilemma: Excessive truncation causes significant portions of responses to be generated by stale model weights, ultimately producing outdated reward signals; conservative truncation diminishes its effectiveness in reducing rollout bubbles.

\myparagraph{Fully asynchronous rollout.} 
AReaL~\cite{fu2025areallargescaleasynchronousreinforcement} adopts a radical approach (\autoref{fig:related-c}):
(1) Fully async rollout. Rollout workers continuously generate responses while train workers select usable samples based on the maximum off-policyness threshold.
(2) Aggressive dependency relaxation. Training can utilize rewards derived from rollouts generated by (potentially) stale model weights many steps prior.
When new model weights are produced, ongoing rollouts are truncated to propagate updated weights. The truncated rollouts are continued after recomputing the KV cache with new weights.
This method has critical limitations:
(1) Fully async rollout changes the paradigm of RL. Excessive off-policyness floods training with obsolete signals.
\noArxiv{Its generalizability to rapidly evolving RL algorithms has not been fully verified.}
(2) Rollout imbalance-induced GPU underutilization remains. Rollouts exceeding the off-policyness threshold are discarded, and frequent KV cache recomputation wastes GPU resources.

\myparagraph{Summary.} In existing systems, rollout imbalance remains a persistent bottleneck, resulting in significant GPU resource underutilization.
Critically, no existing approach reduces the time required for rollout execution and improves the GPU computational underutilization of the rollout stages.

\section{\sys Overview}
To resolve the persistent challenges in LLM RL: time-consuming rollouts and rollout imbalance, 
we design and implement \sys, which innovatively leverages historical rollout information to accelerate rollout execution and minimize imbalance, 
thereby achieving significant speedups in RL training. 

\subsection{Observations and Insights}

\myparagraph{Observation.} 
A complete LLM RL training typically spans 50-100 \emph{epochs}~\cite{shao2024deepseekmathpushinglimitsmathematical,yu2025dapoopensourcellmreinforcement,deepscaler2025}, each comprising multiple \emph{steps} that iterate through the full dataset. 
In the long-term practice of LLM RL training,
we observe strong \emph{\textbf{historical similarity}} in rollout responses across epochs, characterized by:
\begin{itemize}[itemindent=3mm, topsep=1pt, itemsep=0pt, leftmargin=0em,labelindent=0em,style=nextline]
    \item High \emph{token similarity} in rollout responses generated by the same prompt across adjacent epochs, i.e., responses generated in each epoch contain a large proportion of consecutive token sequences that are identical to the sequences in the previous epoch (\autoref{sec:spec-observ});
	\item High \emph{response length distribution similarity} across adjacent epochs, i.e., if we rank the prompts using their response lengths, the ranking order across adjacent epochs remains similar (\autoref{sec:pipe-observ}).
\end{itemize}
The \textbf{root cause} of historical similarity is that current RL algorithms (e.g., PPO\,\cite{schulman2017proximalpolicyoptimizationalgorithms}, GRPO\,\cite{shao2024deepseekmathpushinglimitsmathematical}, DAPO\,\cite{yu2025dapoopensourcellmreinforcement} and GSPO\,\cite{zheng2025groupsequencepolicyoptimization}) apply \emph{clipping operations}~\cite{huang2024ppoclipattainsglobaloptimality} to restrict the magnitude of the model's updates, which maintains stable model evolution. 

\myparagraph{Insights.}
The key insight of \sys is motivated by the above novel observations. 
By systematically organizing and utilizing historical information, which has been overlooked in all existing RL systems,
we unlock new opportunities to further enhance the overall performance of RL system. 
Specifically, to accelerate the rollout process, we propose a dedicated speculative decoding mechanism that leverages historical rollouts as accurate drafts. 
Furthermore, to balance the workload among rollout workers, we introduce a two-tier, distribution-aware pipeline scheduling strategy that exploits the distributional similarity present in historical rollouts.

\begin{figure*}[t]
    \centering
    \includegraphics[width=\linewidth]{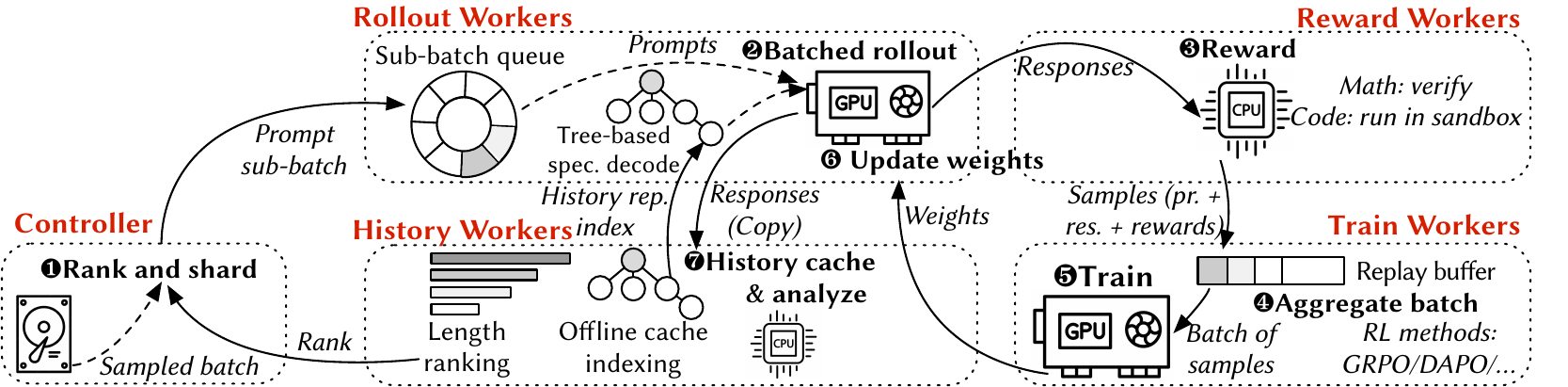}
    \setlength{\belowcaptionskip}{-5pt}
	\setlength{\abovecaptionskip}{-5pt}
    \caption{\textbf{\sys overview.}
    \emph{Solid lines indicate data flow across workers. Dashed lines indicate data dependencies within a worker.}
    }
    \label{fig:design-overview}
\end{figure*}

\subsection{System Overview}
\label{sec:system-overview}

As \autoref{fig:design-overview} shows, \sys inherits the hybrid controller architecture proposed by HybridFlow~\cite{10.1145/3689031.3696075} 
and the disaggregated rollout-train structure (\autoref{fig:related-b}), utilizing dedicated \emph{rollout workers} for response generation, \emph{reward workers} for reward computation, and \emph{train workers} for policy optimization to form a pipelined RL workflow. 

To improve the RL system's overall efficiency, we adopt a \emph{streaming pipeline} architecture.
During each step, each rollout worker retrieves a sub-batch of prompts asynchronously dispatched by the controller from its \emph{prompt sub-batch queue}, 
and generates responses using the inference engine (\ding{183}).
Completed rollout responses (i.e., samples) proceed to reward workers for scoring (\ding{184}) before being transferred to train workers' replay buffer. 
Once the replay buffer accumulates sufficient samples matching the batch size (\ding{185}), the train workers execute user-defined RL algorithms to optimize the model (\ding{186}).
Following full-batch policy optimization, updated model weights are propagated from train workers to the \emph{weight buffers} (on host memory) of rollout workers.
Before processing each step's sub-batch, rollout workers synchronize the weights to their GPUs (\ding{187}).
This \emph{worker-controlled weight update} strategy removes global synchronization overhead and minimizes idle waiting times.

To accelerate rollout generation via speculative decoding (\histospec), \sys employs a \emph{reward-aware suffix-tree-based approach} (\autoref{sub:design:spec-tree}) that efficiently generates speculation drafts from historical data with minimal overhead. 
We further propose an \emph{AIMD-inspired token speculation strategy} (\autoref{sub:design:AIMD}) to dynamically adjust the length of speculative tokens, thereby achieving higher acceptance rates while improving computational density.
Moreover, to balance the workload among rollout workers, \sys implements a \emph{distribution-aware scheduling strategy} (\histopipe), which leverages inter-step complementarity to achieve overall workload balance. It addresses anomalous outliers with \emph{migration-based rebalancing} (\autoref{sub:design:migration-truncation}), and tackles highly skewed rollout time distributions with a \emph{two-tier scheduling} mechanism (\autoref{sub:design:adaptive-parallelism}).

To efficiently manage historical rollout information, we introduce \emph{history workers} that operate on idle CPU resources within RL cluster. 
With the integration of history workers, the RL workflow incorporates two additional stages.
First, the controller leverages the historical length ranking provided by history workers to enable more effective task scheduling (\ding{182}), 
while it also distributes relevant historical responses to the assigned rollout workers. 
Second, after a rollout worker generates responses, it asynchronously sends them to history workers (\ding{188}), 
which update corresponding data structures.

\section{\hspace{-2.4pt}\histospec: Speculative Rollout Generation}
\label{sec:histospec}

Since the rollout phase constitutes the majority of execution time in RL workflows, 
accelerating rollout is critical for improving overall system efficiency. 
Although existing RL systems utilize SOTA inference engines~\cite{10.1145/3600006.3613165, NEURIPS2024_724be447} during rollout, their performance is fundamentally constrained by memory bandwidth, 
due to the extremely long rollout sequences. 
However, we observe that rollout is a specialized LLM inference that demonstrates high historical similarity. 
Motivated by this, we introduce a novel speculative strategy dedicated to the rollout phase, and achieve significant performance improvement in the RL training scenario.

\subsection{Observation: Token Similarity in Rollout}
\label{sec:spec-observ}

RL training comprises multiple iterative epochs (typically 50---100~\cite{shao2024deepseekmathpushinglimitsmathematical}), 
during which the same prompt will be repeatedly sampled across different epochs. 
Mainstream RL algorithms~\cite{schulman2017proximalpolicyoptimizationalgorithms,shao2024deepseekmathpushinglimitsmathematical,yu2025dapoopensourcellmreinforcement,zheng2025groupsequencepolicyoptimization} use clipping operations~\cite{huang2024ppoclipattainsglobaloptimality} to restrict the magnitude of the model's updates.
Furthermore, they use \emph{Group Relative Advantage (GRA) Optimization}, generating multiple responses per prompt (8---64 responses), which enables comprehensive exploration of diverse reasoning trajectories throughout each epoch.
Consequently, the outputs generated for the same prompt during rollouts at adjacent epochs exhibit a high degree of similarity.

\begin{figure*}[t]
    \centering
    \begin{subfigure}[b]{0.33\textwidth}
        \centering
        \includegraphics[width=\textwidth]{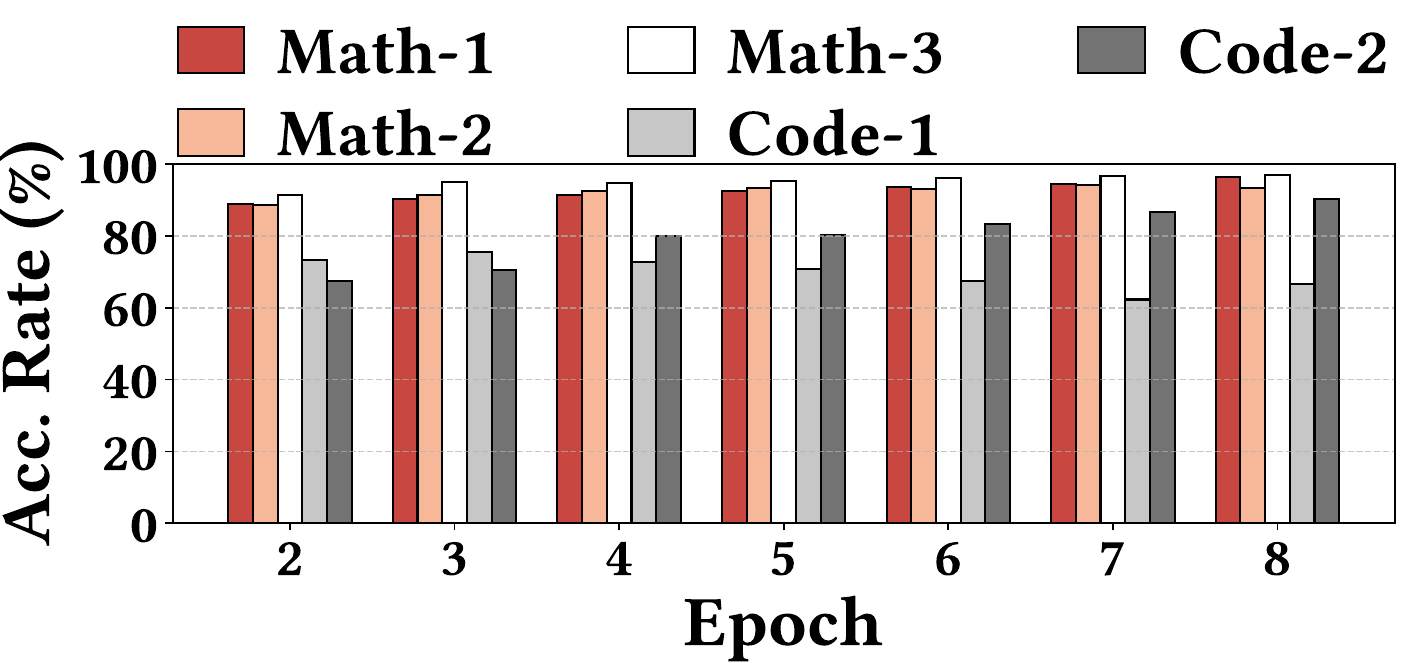}
        \caption{\textbf{Acceptance rate across epochs}}
        \label{fig:spec-hit-rate-overall}
    \end{subfigure}
    \hfill
    \begin{subfigure}[b]{0.52\textwidth}
        \centering
        \includegraphics[width=\textwidth]{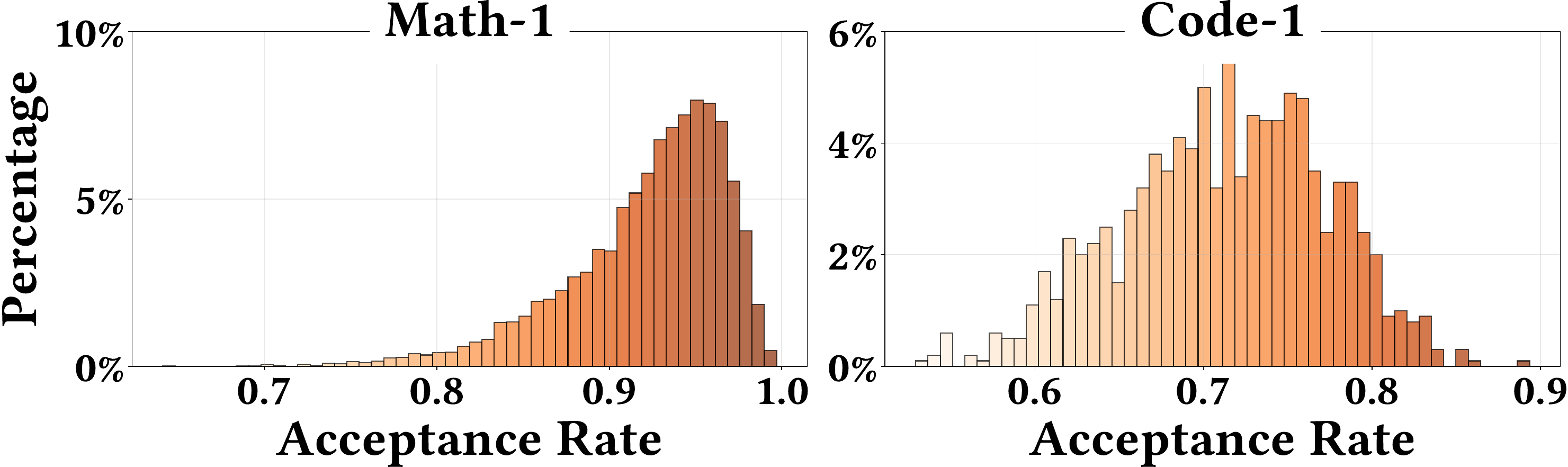}
        \caption{\textbf{Responses' acceptance rate distribution (Math-1 and Code-1)}}
        \label{fig:spec-hit-rate-math}
    \end{subfigure}
    \hfill
    \begin{subfigure}[b]{0.13\textwidth}
        \centering
        \fontsize{8.5pt}{11.5pt}\selectfont
        \begin{tabular}{l|c}
            \cline{1-2}
            \textbf{Dataset} & \textbf{Size} \\
            \cline{1-2}
            \textbf{Math-1} & 230K \\
            \cline{1-2}
            \textbf{Math-2} & 40K \\
            \cline{1-2}
            \textbf{Math-3} & 10K \\
            \cline{1-2}
            \textbf{Code-1} & 40K \\
            \cline{1-2}
            \textbf{Code-2} & 12K \\
            \cline{1-2}
        \end{tabular}
	    \setlength{\abovecaptionskip}{10pt}
        \caption{\textbf{Datasets used}}
        \label{fig:datasets}
    \end{subfigure}
    \setlength{\belowcaptionskip}{-5pt}
	\setlength{\abovecaptionskip}{2pt}
    \caption{\textbf{Characterizing historical token similarity.}
        \emph{We use five datasets to train 14B LLMs respectively using the GRPO algorithm.}
    }
    \label{fig:spec-hit-rate}
\end{figure*}

To quantitatively evaluate the similarity of tokens generated during rollouts, 
we analyze the response traces generated during RL training across five math/code datasets, as listed in \autoref{fig:datasets}.
For each response, we simulate its rollout ``generation'' from the beginning, and use the last three ``generated'' tokens as the prefix to search for exact matches in the historical responses from the previous epoch. 
When a match is found, we record the length of identical token sequences that follow the prefix in both historical and current responses, labeling them as ``accept'' tokens.
Otherwise, we will forward the response to ``generate'' one token.
The routine is repeated until the end of the response.
As \autoref{fig:spec-hit-rate-overall} shows, across 8 epochs, for math tasks, 93\% of tokens could be successfully ``accept'' using this routine (on average, 75\% for code). 
Furthermore, as training progresses (i.e., with increasing epochs), the similarity increases.
\autoref{fig:spec-hit-rate-math} shows the distribution of the responses' acceptance rates.

\subsection{Speculative Rollout with History}
Leveraging historical token similarity, we design \histospec, which utilizes speculative decoding to accelerate rollouts and uses historical responses as draft sources. 
In each decoding iteration, \histospec uses \emph{the last few generated tokens} as the prefix to search for matches within the prompt's historical responses.
Upon matching, it extracts a certain number of subsequent tokens following the prefix as drafts.
The drafts are then verified via a single LLM forward pass, with some (possibly all) tokens accepted (\autoref{sec:spec-back}).
This history-based speculative rollout improves computational resource utilization during rollouts, and reduces the time of the rollout stage.

\myparagraph{Technical challenges.} However, this approach encounters two challenges: (1) Over-long rollout sequences induce significant prefix matching overhead, and the branching in historical responses complicates draft token selection;
(2) Unpredictable draft acceptance length creates a trade-off between underutilizing compute capacity (when predicting too few tokens) and wasting resources on verifying rejected tokens (when over-predicting).

\begin{figure}[t]
    \centering
    \includegraphics[width=\linewidth]{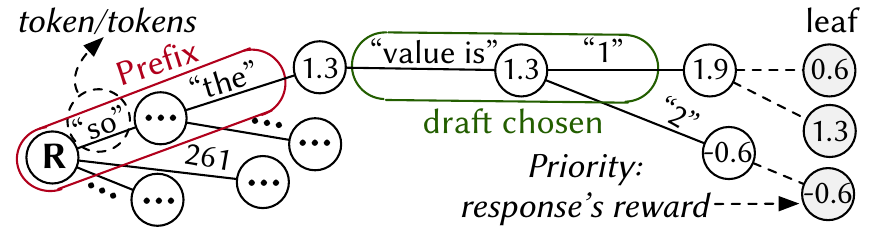}
    \setlength{\belowcaptionskip}{-5pt}
	\setlength{\abovecaptionskip}{-5pt}
    \caption{\textbf{Suffix-tree based draft generation.}
    \emph{R\,=\,Root. For ease of understanding, we convert token ids into words.
    }
    }
    \label{fig:suffix}
\end{figure}

\subsection{Tree-based Historical Rollout Management}
\label{sub:design:spec-tree}
Draft generation overhead is a pivotal factor determining speculative decoding's effectiveness. 
However, existing corpus-based speculation methods face a dilemma between retrieval costs and index building costs. E.g., n-gram~\cite{saxena2023prompt, vllm-spec} requires time-consuming linear scans for prefix matching, and SuffixDecoding~\cite{oliaro2025suffixdecodingextremespeculativedecoding} limits the lengths of draft source sequences to reduce index building costs.
During LLM RL, each epoch generates multiple long responses (totaling hundreds of thousands of tokens per prompt). Traditional draft generation methods struggle to operate efficiently under the constraints.

\myparagraph{Async cache building.}
The RL workloads allow us to relax the constraints on draft generation. 
In RL sampling strategies, the same request is generally not re-sampled until multiple steps later.
Therefore, we are able to shift the computational overhead associated with retrieval to the indexing phase.
\sys employs history workers to index historical rollout sequences asynchronously.
Upon generating new responses, the controller dispatches them to history workers for index construction.
When scheduling prompts to rollout workers (asynchronously), the controller notifies the corresponding history workers to transfer the relevant indexed cache.

\myparagraph{Reward-aware tree-based management.}
\sys employs suffix trees~\cite{10.1109/SWAT.1973.13} to index cached responses, which enables $O(m)$-time matching for length-$m$ prefixes.
\sys maintains a dedicated tree for each prompt, indexing all its historical responses generated in its last rollout.
Since modern RL algorithms generate multiple responses per prompt to explore multiple solution paths, a prefix may branch to divergent suffixes, complicating draft token selection.
We observe that, during training, RL algorithms optimize the model towards generating high-reward solutions with higher probability.
Therefore, we add a \emph{priority} value to each tree node, weighted by the sum of the rewards of its branch.
For every suffix branch present, \histospec selects the one with the highest priority.
Through this RL-algorithm-guided design, \histospec maximizes the speculation acceptance rate.

As \autoref{fig:suffix} shows, in the suffix tree, each node represents a subsequence formed by the path from root to that node, where each leaf node corresponds to a complete suffix of a response, and each edge represents one or more tokens extending the sequence of parent node to the sequence of the child.
For each leaf node, its priority (marked on the nodes in \autoref{fig:suffix}) is the sum of the rewards for the responses that end with this suffix (that the leaf node represents).
A parent node's priority is the sum of its children's priorities.

\myparagraph{Resources.} Both the tree's construction time overhead and memory overhead are $O(n)$ for $n$ tokens~\cite{10.1007/BF01206331}. 
GPU clusters have sufficient CPU resources (64--128+ cores and multi-TB host memory), which are mostly idle during previous RL flows. \sys strictly limits the resources used by history workers using OS methods~\cite{cgroups}, preventing them from interfering with other workers. For RL training with a 230K dataset and 16K max response length, the host memory overhead of suffix trees is $<$\,80GB per node with 8 nodes.
\histospec also supports compression and swapping to SSD for memory saving, and checkpoint for fault tolerance.

\subsection{AIMD-like Token Speculation}
\label{sub:design:AIMD}

\begin{figure}[t]
    \centering
    \includegraphics[width=0.85\linewidth]{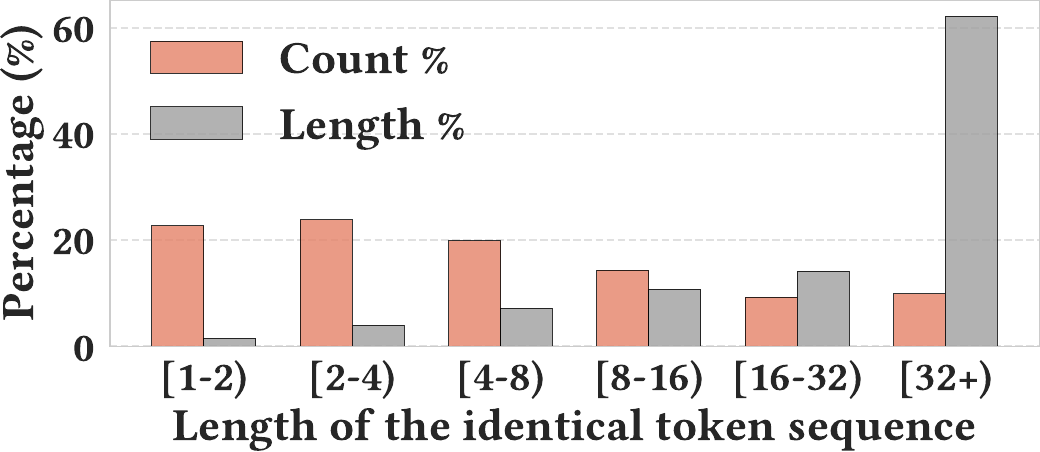}
    \setlength{\belowcaptionskip}{-5pt}
	\setlength{\abovecaptionskip}{5pt}
    \caption{\textbf{Distribution of speculative hit lengths.}
    }
    \label{fig:spec-length-count}
\end{figure}

To determine the length of tokens predicted by \histospec in each iteration,
we conduct a detailed analysis of the length distribution of identical token sequences in rollout responses.
As shown in \autoref{fig:spec-length-count}, short segments (1-2 tokens) dominate in quantity, while long segments account for the majority in the total length.
This poses a challenge for \histospec:
If we predict many tokens per iteration, most tokens would be rejected, wasting significant computation on verifying them.
If we predict a small number of tokens, computational resources cannot be fully utilized, undermining speculative decoding's advantage in improving computational density.

We find that network congestion control encounters similar challenges. Classical TCP congestion control employs the AIMD (\emph{Additive Increase, Multiplicative Decrease}~\cite{896303}) principle:
gradually increasing the window size when network is uncongested, but aggressively reducing it upon congestion.
Inspired by AIMD, \histospec designs a dynamic speculation window for each response, initialized to two tokens.
When all speculated tokens are accepted, \histospec additively increases the window size by 2, until it meets the upper threshold (32 by default).
But when any token is rejected, it resets the size directly to 2.
This approach guarantees high computational density during generating long matching sequences, and minimal wasted computation for short sequences.

As for the prefix length, \histospec sets it to 7 initially. 
If no matching suffix is found, \histospec progressively reduces it until it reaches 3.
These hyperparameters are adjustable.

\histospec also considers \emph{batch size}. At large batch sizes, increased decoding parallelism yields higher GPU computational utilization, where excessively low speculative acceptance rates degrades the overall throughput.
To mitigate this, \histospec monitors the acceptance rates and adaptively disables speculation.
Through pre-profiling, \histospec gets the maximum viable batch size for throughput gains at different acceptance rates. 
When a rollout worker's current batch size exceeds the threshold, speculation is automatically disabled to preserve system efficiency.

\section{\hspace{-2.4pt}\histopipe: Hybrid Rollout Pipeline}
\label{sec:design:histopipe}

Although \histospec improves rollout efficiency, it does not address the issue of imbalanced rollout distributions, 
resulting in non-trivial GPU resource underutilization. 
We observe that leveraging the information from historical rollouts can effectively maintain load balance throughout 
the rollout process,
and propose the \histopipe scheduling design.

\begin{figure}[t]
    \centering
    \includegraphics[width=\linewidth]{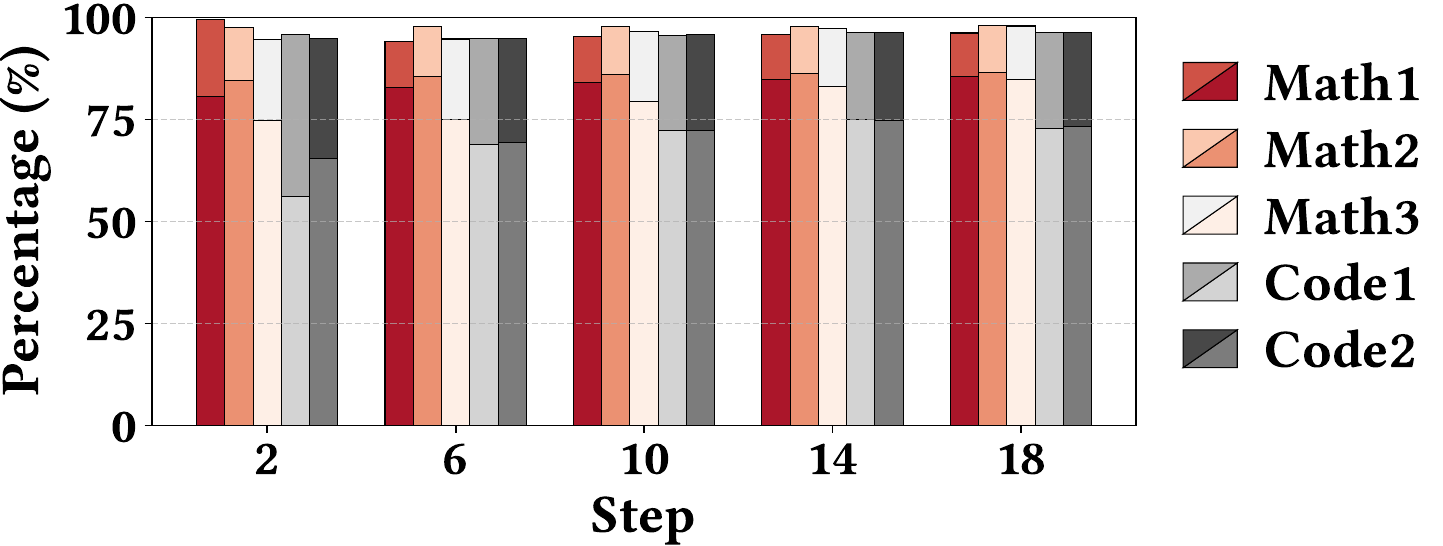}
    \setlength{\belowcaptionskip}{-5pt}
	\setlength{\abovecaptionskip}{-5pt}
    \caption{\textbf{The ranking group changes of responses across 20 epochs.}
    \emph{Since RL algorithms like GRPO generate multiple responses for each prompt, we analyze the previous epoch by assigning all responses of each prompt to the same ranking group based on their \textbf{median} length, yielding the \underline{predicted group}. 
    For the current epoch, we sort all responses by their actual lengths to assign groups, obtaining the \underline{real group}, and compare it with the predicted group to collect the data.
    In the figure, the lower parts of the bars represent the proportion of responses that remain in the same or a lower group; the upper parts represent the proportion of responses that, despite changing groups, only move up one group and have lengths in the shorter 50\% of the new group (i.e., shift near the group boundary).}
    }
    \label{fig:motiv-rank}
\end{figure}

\subsection{Observation: Distribution Similarity in Rollout}
\label{sec:pipe-observ}

Although different prompts generate varying numbers of tokens within a single rollout, 
we observe that the response length (i.e., token count) distribution for the same prompt across adjacent epochs remains similar. 
In other words, if a prompt generates a relatively large number of tokens in one rollout iteration, 
it is likely to produce a similarly large token count in subsequent rollouts.
Therefore, \emph{we can effectively predict the ranking of response lengths using historical lengths.}

To validate this observation, we analyze the response length rankings across multiple rollout epochs for five datasets (\autoref{fig:datasets}).
Specifically, we group the responses into 8 ranking groups based on their lengths, ranked from low to high. 
As \autoref{fig:motiv-rank} shows, across 20 epochs (300---1000 steps), for math tasks, an average of only 16\% of responses change their group to higher ones (28\% for code tasks).
Among them, 13\% of the (all) responses only shift near the group boundary without significantly altering the distribution (24\% for code).
These results demonstrate that not only do generated tokens exhibit similarity across rollouts, 
but the \emph{distribution of rollout lengths also remains relatively consistent over epochs}.

\subsection{Distribution-aware Hybrid Pipeline}
\label{sec:hybrid-pipe}

Leveraging the similarity in rollout length distributions can enable more balanced scheduling of rollout tasks, 
reducing load imbalance during rollout. 
We propose a distribution-aware rollout pipeline strategy, named \histopipe.

\myparagraph{Hybrid Pipeline.}
Preserving the rollout-reward-train dependency is essential for algorithmic integrity. 
However, the imbalanced distribution of rollout lengths and the autoregressive nature of LLMs induce rollout bubbles within individual steps. 
Inspired by Dual-Pipe's philosophy~\cite{deepseekai2025deepseekv3technicalreport}, instead of seeking per-step balance, \histopipe shifts focus to \emph{inter-step workload complementarity},
which constructs synergistic balancing across consecutive training steps.

\begin{figure}[t]
    \centering
    \includegraphics[width=\linewidth]{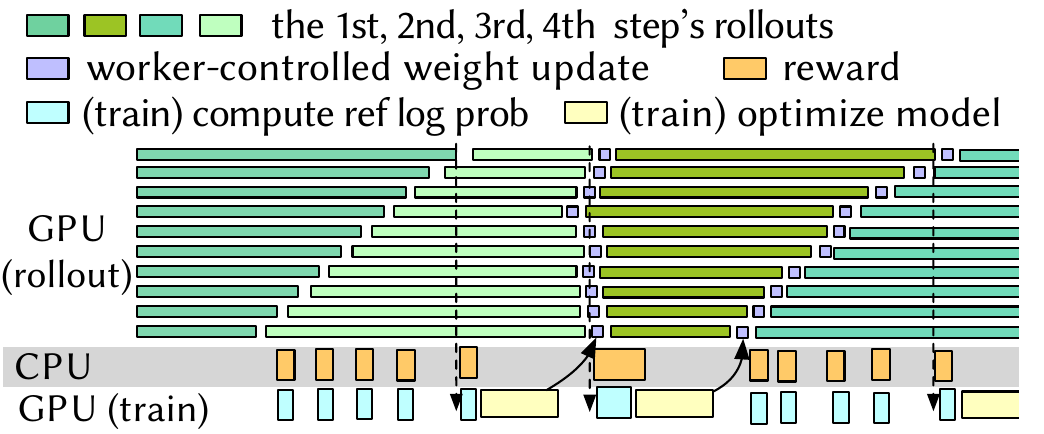}
    \setlength{\belowcaptionskip}{-5pt}
	\setlength{\abovecaptionskip}{-10pt}
    \caption{\textbf{\histopipe design.}
    \emph{
    Leveraging the \textbf{worker-controlled async weight updating} (\autoref{sec:system-overview}), updated model weights are propagated asynchronously to the 
    rollout workers' weight buffers and updated to GPUs by rollout workers upon completing a rollout step.
    In rare cases where the corresponding weights have not been propagated to the weight buffer, the rollout worker will await the weights.
    }
    }
    \label{fig:design-histopipe}
\end{figure}

Historical distribution enables \histopipe to rank rollout prompts based on their historical response lengths, obtaining several \textbf{ranking groups} by equally dividing the ranked prompts.
As \autoref{fig:design-histopipe} shows, during odd-numbered rollout steps, the scheduler assigns ranking groups to workers (0 through \emph{N-1}) in \emph{ascending order} of rollout length.
Conversely, during even-numbered steps, ranking groups are assigned in \emph{descending order} of rollout length.
By complementing rollout lengths in this alternating manner, 
we effectively fill idle times or bubbles that occur during rollout execution, thereby improving the overall efficiency of the rollout workers.

\histopipe is not enabled during the first epoch as no historical information exists, which is acceptable because
RL training typically spans 50---100 epochs, and the first epoch exhibits the shortest duration.
For RL algorithms using Group Relative Advantage, \histopipe uses the median lengths within each response group as the \emph{historical response lengths} to rank the prompts.
Although we do not preclude more complicated algorithms or model-based methods~\cite{zhong2025streamrlscalableheterogeneouselastic},
the current method is sufficiently robust and effective (\autoref{fig:motiv-rank}). 

\myparagraph{Technical challenges.}
\histopipe also faces several challenges under real workloads.
(1) Although the rollout length distributions exhibit high similarity, 
the occurrence of an anomalously long rollout can damage the complementarity and disrupt the pipelines.
(2) In practice, the long-tail distribution of rollout lengths prevents consecutive steps from achieving perfect workload complementarity.

\subsection{Migration-based Rebalancing}
\label{sub:design:migration-truncation}

To mitigate the impact of anomalous rollout lengths on the hybrid pipeline, we employ two strategies. 
(1) \emph{Intra-step migration.} Migrating excessively long rollouts to other groups that are still in the same step's rollout process.
(2) \emph{Inter-step migration.} Migrating the outliers to the next step.

More specifically, we set a threshold for each ranking group. When a rollout length exceeds the threshold, it will be migrated. 
For the first strategy, long rollouts are dynamically reassigned to other groups during execution.
The generated tokens are preserved and the rollouts continue after \sys recomputes the KV cache (i.e., prefill) of the prompt and the tokens. \footnote{We do not migrate the KV cache because the overall overhead of KV cache migration exceeds that of recomputation.}
For the second strategy, the migrated rollout is added to the next step.
Similarly, the generated tokens are preserved and the rollouts continue after KV cache recomputation.
Since the RL algorithms and systems inherently use oversampling, 
inter-step migrating a small number of rollouts and deferring their completion to the next step do not adversely affect training.
In our scheduling design, anomalous rollouts in groups with short rollout lengths are preferentially intra-step migrated, 
whereas inter-step migration is used for anomalous rollouts in groups with generally long rollout lengths.
In practice, intra-step migration efficiently handles most of the outliers.

\myparagraph{Threshold.}
\sys migrates rollouts exhibiting:
(1) within the last remaining $\alpha\%$ of the rollouts in the group;
(2) generated response length exceeding $\beta$-times the maximum historical response length of the prompts in the same group.
Currently, $\alpha$=10 and $\beta$ is determined by the 75th percentile of the growth rates in response lengths of the previous epoch ($\beta$=1.1 if the percentile <\,1.1).
Analysis across 5 datasets over 20 epochs reveals only 1.49\%---3.55\% (8 groups, on average) of responses undergo migration, demonstrating acceptable overhead and negligible impact on training accuracy.

\begin{table}[t]
    \centering
    \setlength\tabcolsep{0.4\tabcolsep}
	\fontsize{8pt}{9.6pt}\selectfont
    \begin{tabular}{l|cc|cc|cc|cc|cc}
        \cline{1-11}
        \textbf{Dataset name} & \multicolumn{2}{c|}{\textbf{Math-1}} & \multicolumn{2}{c|}{\textbf{Math-2}} & \multicolumn{2}{c|}{\textbf{Math-3}} & \multicolumn{2}{c|}{\textbf{Code-1}} & \multicolumn{2}{c}{\textbf{Code-2}} \\
        \cline{1-11}
        \textbf{Rank}           & 8            & 16            & 8         & 16        & 8          & 16        & 8         & 16        & 8         & 16 \\
        \textbf{Accurate}       & 85.7          & 79.7    & 83.2         & 76.9          & 79.0         & 71.6         & 71.8      & 62.6         & 73.2         & 62.5\\
        \textbf{Not last 10\%}  & 12.2          & 16.8    & 12.7         & 17.5           & 16.9         & 22.8         & 23.9      & 30.4         & 22.9         & 30.1\\
        \textbf{Within 1.1x}  & 0.61          & 1.03    & 0.55         & 1.02           & 0.86         & 1.47         & 1.44      & 2.34         & 1.46         & 2.32\\
        \textbf{Migrated}      & \textbf{1.49} & \textbf{2.47}   &\textbf{3.55}& \textbf{4.58}  & \textbf{3.24}& \textbf{4.13}&\textbf{2.86}&\textbf{4.66}& \textbf{2.44}& \textbf{5.08}\\
        \cline{1-11}
    \end{tabular} \\[0pt]
    \setlength{\belowcaptionskip}{0pt}
    \setlength{\abovecaptionskip}{0pt}
    \caption{\textbf{Accuracy of predicting the samples' ranks with historical lengths and migration rates (14B models).}
    \emph{
        We list the average value of the metrics of the 20+ epochs.
        }
    }
    \label{tab:design-rank-all-dataset}
\end{table}

\subsection{Two-tier Scheduling}
\label{sub:design:adaptive-parallelism}

\begin{figure}[t]
    \centering
    \includegraphics[width=\linewidth]{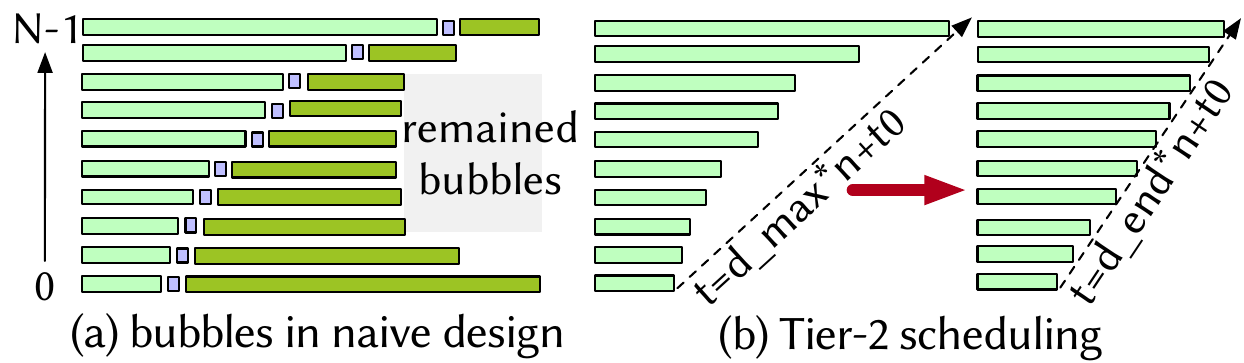}
    \setlength{\belowcaptionskip}{-5pt}
	\setlength{\abovecaptionskip}{-8pt}
    \caption{\textbf{Bubbles caused by skewed rollout time distribution, and how the two-tier scheduling reduces bubbles.}
    \emph{}
    }
    \label{fig:design-two-tier}
\end{figure}

Achieving near-bubble-free inter-step complementarity requires approximately linear execution time distributions across rollout groups.
In practice, however, long-tailed rollouts (\autoref{fig:response-len-distri}) cause the time distributions to approximate exponential patterns, which leads to bubbles on some rollout workers, as \autoref{fig:design-two-tier}a shows.
\histopipe proposes a two-tier scheduling mechanism to tackle this issue:
\begin{itemize}[itemindent=3mm, topsep=1pt, itemsep=0pt, leftmargin=0em,labelindent=0em,style=nextline]
    \item \textbf{Tier-1: from prompts to ranking groups.} First, \histopipe equally divides the ranked prompts into ranking groups.
	\item \textbf{Tier-2: mapping ranking groups to GPUs.} If we allocate equal GPUs per ranking group, their execution times will exhibit exponential distributions due to long-tail rollouts. Therefore, \histopipe designs a distribution reshaping strategy: instead of evenly distributing GPUs among the groups, it 
    allocates fewer GPUs to the short-length and medium-length groups while provisioning extra GPUs to the few groups with long responses. In this way, \histopipe reshapes the groups' rollout time distribution from \emph{exponential to linear}, thereby further reducing rollout bubbles. 

\end{itemize}

In \sys, each rollout worker manages several GPUs cooperating via model parallelism with user-specified configurations. 
During Tier-2 scheduling, \sys allocates GPUs at rollout-worker granularity. 
Rollout workers operate via data parallelism, thus, increasing rollout workers reduces per-worker batch size and shortens the execution time.

\begin{figure}[t]
	\setlength{\belowcaptionskip}{-5pt}
	\setlength{\abovecaptionskip}{3pt}
	\centering
	\setlength{\columnsep}{0.2cm}
	\begin{minipage}[t]{0.49\linewidth}
		\begin{minted}[
    numbersep=\parindent,
    escapeinside=||,
    breaklines,      %
    fontsize=\fontsize{8pt}{9pt}\selectfont, %
    tabsize=0        %
    ]{python}
|\Ddhighlight{MIN\_WKS/MAX\_WKS: a ranking}|
|\Ddhighlightnotri{group\textquotesingle\, min/max worker number}|
|\Ddhighlight{Tool func: calculate workers}|
|\Ddhighlightnotri{needed for a given d}|
|\highlightfuncword{func}| |\Ddhighlightfuncname{calWks}|(d, lens, wks, t0)
  wks_needed = 0 
  for i in range(0, N):
    target_t = t0+i*d
    |\Ddhighlightcomment{find min workers to meet}|
    |\Ddhighlightcommentnotri{the group\textquotesingle\,s target time}|
    group_wks = -1
    for k in range(MIN_WKS, 
            MAX_WKS+1):
      exec_t = |\Ddhighlightfunc{$\tau$}|(lens[i], k)
      if exec_t <= target_t:
        group_wks = k
        break
    if group_wks == -1:
      return (Inf, [])
    wks_needed += group_wks
    plan[i] = group_wks
  return (wks_needed, plan)
\end{minted}
	\end{minipage}\hfill%
	\begin{minipage}[t]{0.49\linewidth}
		\begin{minted}[
    numbersep=\parindent,
    escapeinside=||,
    breaklines,      %
    fontsize=\fontsize{8pt}{9pt}\selectfont, %
    tabsize=0        %
    ]{python}
|\Ddhighlight{Find best worker allocation}|
|\Ddhighlight{The target time of group n:}|
|\Ddhighlightnotri{t(n)=d*n+t0, 0\,<=\,n\,<\,N}|
|\highlightfuncword{func}| |\Ddhighlightfuncname{planAllocation}|(lens,  
    wks, t_train)
 t0 = max(|\Ddhighlightfunc{$\tau$}|(lens[0], MAX_WKS), 
     t_train)
 d_min = 0.0
 d_max = (|\Ddhighlightfunc{$\tau$}|(lens[N-1],
        MIN_WKS)-t0)/(N-1)
 p = 1 # precision, adjustable
 |\Ddhighlightcomment{binary search}|
 while (d_max-d_min) > p:
  d_mid = (d_max+d_min)/2
  result = |\Ddhighlightfunc{calWks}|(d_mid, 
            lens, wks, t0)
  if result.wks_needed > wks:
   d_min = d_mid
  else: # feasible solution
   best_plan = result.plan
   d_max = d_mid
 return best_plan

\end{minted}
	\end{minipage}
	\caption{\textbf{Tier-2 scheduling algorithm.}
    \emph{
        It solves the problem by minimizing the execution time gradient ($d$) via binary search, as \autoref{fig:design-two-tier}b shows.
        We use \codeword{t\_train} (time of the last finished step's training stage) to ensure the execution time of the shortest ranking group is longer than the training stage's time, avoiding the next step's rollout waiting for train workers to generate weights.
    $\tau(l, dp)$ is determined by looking up the table obtained from pre-profiling.
    The \textbf{cooperation of \histopipe with \histospec} is also considered. $\tau(l, dp)$ also considers \histospec's current speculation information (e.g., acceptance rate), which is omitted in the algorithm for ease of understanding.
}
    }
	\label{fig:algorithm}
\end{figure}

\myparagraph{Algorithm.}
\histopipe determines the GPU allocation plan using the algorithm detailed in \autoref{fig:algorithm}.
It uses the mean value of the prompts' historical response lengths within each ranking group (\emph{representative length}) to estimate the group's execution time.
Through pre-profiling of the inference engine (i.e., \histospec, also considering the impact of speculative decoding), we can obtain the relationship between a ranking group's execution time \emph{(t)}, its representative rollout length \emph{(l)}, and DP worker number \emph{(dp)}: $t$=$\tau(l, dp)$.
Within the RL cluster, there are $wks$ rollout workers to serve $N$ ranking groups, whose representative lengths are $lens$.
We have to assign varying numbers of rollout workers to each group, and make their completion times close to linear distribution.
This constitutes an integer nonlinear programming problem and we converge to the solution via binary search.

\begin{figure*}[t]
    \centering
    \includegraphics[width=\linewidth]{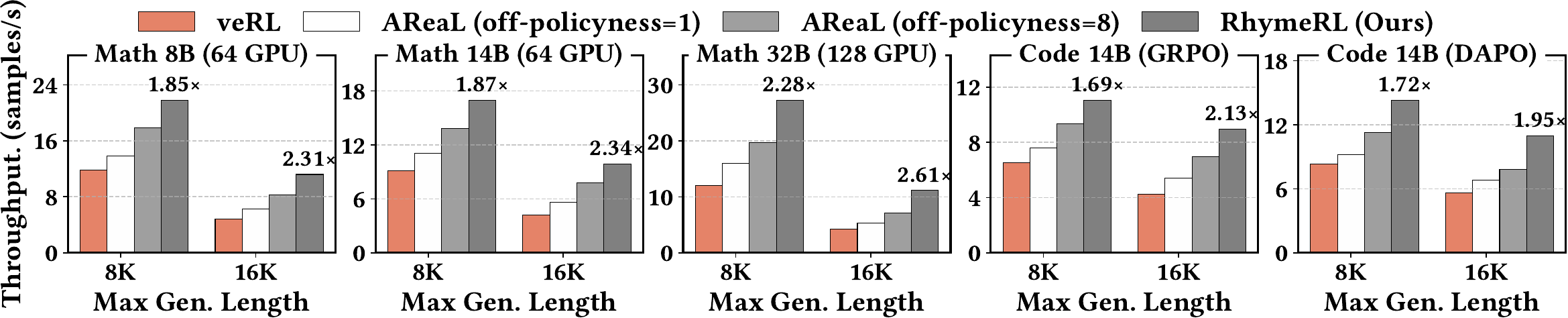}
    \setlength{\belowcaptionskip}{-5pt}
	\setlength{\abovecaptionskip}{-7pt}
    \caption{\textbf{The training throughput of \sys, veRL, and AReaL.} 
    \emph{For each setting, we use the same configurations (e.g., number of rollout/train workers, clip ratio, etc.) for the three systems.
    We use 4-GPU tensor parallelism (TP) for 32B LLMs, and 2-GPU TP for 8B/14B LLMs.
     }
    }
    \label{fig:eval-tp}
\end{figure*}

\section{Evaluation}
\label{s:eval}

We evaluate \sys with LLMs of sizes ranging from 8B to 32B, on real-world math and code datasets.

\myparagraph{Experiments.}
First, we compare \sys's end-to-end training throughput with SOTA LLM RL systems on training LLMs of different sizes using different datasets (\autoref{sec:eval-end-to-end}).
Then, we break down the improvements of \sys's designs (\autoref{sec:eval-ablation}).
To gain a deeper understanding of \histospec (\autoref{sec:eval-spec}), we study \histospec's improvements on rollout throughput across steps, speculation rates, and acceptance rates.
To further understand \histopipe (\autoref{sec:eval-pipe}), we study its improvements on training throughput across steps, and its migration rates.
Finally, we study the impact of \sys on training accuracy and algorithm behavior (\autoref{sec:accuracy}).
We also compare \histospec with other speculative decoding methods (\autoref{sec:compare-with-other}).

\myparagraph{Hardware settings.}
We deploy \sys on a GPU cluster with 16 nodes and 128 \gpuSHORT GPUs.
Each node is equipped with 8 \gpu GPUs, 2 Intel Xeon Sapphire Rapids CPUs with 96 cores, 1900GB host DRAM, and 9 * 400 Gb/s NVIDIA ConnectX7 InfiniBand NIC (8 GPU-dedicated NICs and one CPU-dedicated NIC).

\myparagraph{Model and algorithm.}
We verify \sys's observations and performance improvements on Qwen-2.5 models~\cite{qwen2025qwen25technicalreport}, Qwen-3 models~\cite{yang2025qwen3technicalreport}, DeepSeek-R1-Distill-Qwen models~\cite{deepseekai2025deepseekr1incentivizingreasoningcapability} and LLama-3 models~\cite{grattafiori2024llama3herdmodels}.
Our evaluation utilizes Qwen3-8B-Base, Qwen3-14B-Base, and Qwen2.5-32B models,
with detailed architectures listed in \autoref{tab:model-specs}.
We train these models using GRPO algorithm~\cite{shao2024deepseekmathpushinglimitsmathematical}, which is currently the mainstream LLM RL algorithm, powering advanced LLMs such as DeepSeek-R1~\cite{deepseekai2025deepseekr1incentivizingreasoningcapability}.
We also verify \sys's efficiency on recent next-generation algorithms like DAPO~\cite{yu2025dapoopensourcellmreinforcement} and provide the results.
Based on our practice, the response group size is set to 16, which yields peak algorithmic efficiency.

\begin{table}[t]
    \centering
    \setlength\tabcolsep{0.4\tabcolsep}
    \fontsize{8.5pt}{10pt}\selectfont
    \begin{tabular}{cc|ccccc}
        \cline{1-6}
        \textbf{Model arch} & \textbf{Size} & \textbf{Layers} & \textbf{Attn heads} & \textbf{K/V heads} & \textbf{Hidden size} \\
        \cline{1-6}
        \textbf{Qwen-3} & \textbf{8B} & 36 & 32 & 8 & 4096 \\
        \textbf{Qwen-3} &\textbf{14B} & 40 & 40 & 8 & 5120 \\
        \textbf{Qwen-2.5} &\textbf{32B} & 64 & 40 & 8 & 5120 \\
        \cline{1-6}
    \end{tabular} \\[-10pt]
    \setlength{\belowcaptionskip}{10pt}
	\setlength{\abovecaptionskip}{-5pt}
    \caption{\textbf{The specifications of the evaluated LLMs.}
    }
    \label{tab:model-specs}
\end{table}

\myparagraph{Metrics.} 
As per convention~\cite{park2017hybrid, zhong2025streamrlscalableheterogeneouselastic, 305943}, we report training throughput, measured by the average number of samples generated and processed per second. 
We sample 8K math/code prompts from internal datasets (\autoref{fig:datasets}).
Under each setting, we train 80 steps for math models and 100 steps for code models, and use the first 20 steps as warm-up.

\subsection{End-to-end Training Performance}
\label{sec:eval-end-to-end}

\begin{figure}[t]
    \centering
    \includegraphics[width=\linewidth]{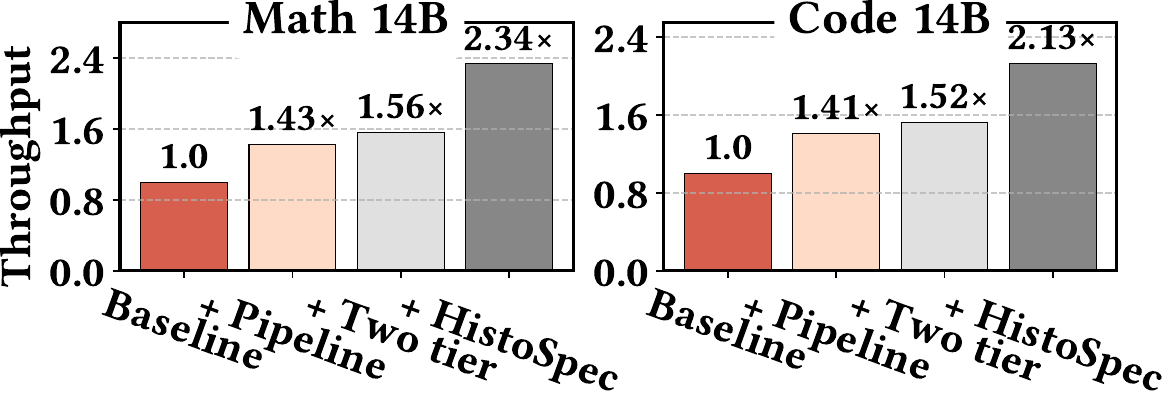}
    \setlength{\belowcaptionskip}{-5pt}
	\setlength{\abovecaptionskip}{-8pt}
    \caption{\textbf{Breakdown of \sys's improvements.}
    \emph{The max response length is set to 16K tokens. The throughputs are normalized.}
    }
    \label{fig:eva-ablation}
\end{figure}

We compare the end-to-end training performance of \sys with the following state-of-the-art LLM RL training systems.
\begin{itemize}[itemindent=3mm, topsep=1pt, itemsep=0pt, leftmargin=0em,labelindent=0em,style=nextline]
    \item \textbf{veRL~\cite{10.1145/3689031.3696075}} (v0.4.1). Featuring the hierarchical hybrid programming model and highly-optimized 3D-HybridEngine, veRL is used by many projects and companies, and is \sys's codebase. We use the rollout/train disaggregated architecture natively supported by veRL, which is the SOTA RL architecture and outperforms its hybrid mode~\cite{verl-one-step-off}.
    \item \textbf{AReaL~\cite{fu2025areallargescaleasynchronousreinforcement}} (v0.3.0). Leveraging fully async rollout, AReaL achieves continuous rollout GPU utilization without idle time (\autoref{sec:existing-work}). 
    We evaluate AReaL's performance when its max off-policyness threshold is 1 (equal off-policyness with \sys and veRL) and 8 (the max off-policyness recommended).
    \end{itemize}

As \autoref{fig:eval-tp} shows, \sys outperforms veRL and AReaL on different model sizes (8B---32B), response lengths (8K/16K), tasks (Math/Code) and algorithms (DAPO/GRPO).
Compared with veRL, \sys improves the training throughput by up to 2.6x (1.9x on average for 8K max response length, and 2.3x on average for 16K max response length). This is primarily attributed to \sys significantly reducing rollout time and effectively minimizing rollout bubbles.
Compared with AReaL, when its off-policyness is 1, \sys improves the training throughput by up to 2.1x (1.6x on average for 8K max response length, and 1.8x on average for 16K max response length).
Although AReaL achieves continuous GPU utilization, it introduces rollout truncation and KV cache recomputation overhead.
Leveraging historical similarity, \sys achieves a more effective rollout pipeline.
When AReaL's off-policyness is 8, \sys improves the training throughput by up to 1.6x (1.3x on average for 8K max response length, and 1.4x on average for 16K max response length).
\sys outperforms AReaL (off-policyness\,=\,8) as it significantly reduces rollout time through \histospec. Besides, \sys does not change current RL paradigm.

\subsection{Extended Studies}
\subsubsection{Ablation Study}
\label{sec:eval-ablation}

We break down the improvements brought by \sys's designs. 
As \autoref{fig:eva-ablation} shows, for the Math-14B LLM, \histopipe (\autoref{sec:hybrid-pipe}) achieves 1.43x (1.41x for Code-14B) throughput boosts, with Two-tier Scheduling (\autoref{sub:design:adaptive-parallelism}) further improving the throughput of the naive hybrid pipeline by 1.10x (1.08x for Code-14B).
\histospec (\autoref{sec:histospec}) further improves the training throughput by 1.50x (1.40x for Code-14B).

\begin{figure}[t]
    \centering
    \includegraphics[width=\linewidth]{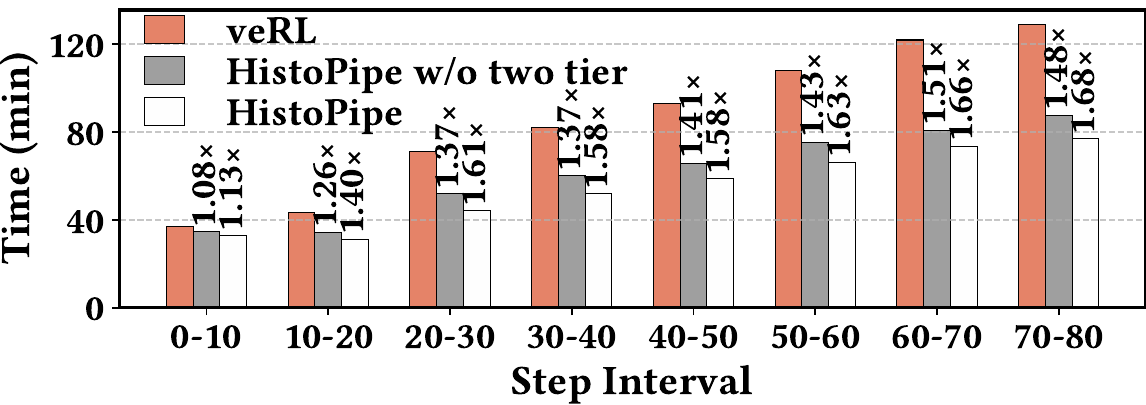}
    \setlength{\belowcaptionskip}{-5pt}
	\setlength{\abovecaptionskip}{-8pt}
    \caption{\textbf{\histopipe's improvements across steps.}
    \emph{Model: Math-14B. Max response length\,=\,16K.}
    }
    \label{fig:eva-pipe}
\end{figure}

\begin{figure*}[t]
    \centering
    \includegraphics[width=\linewidth]{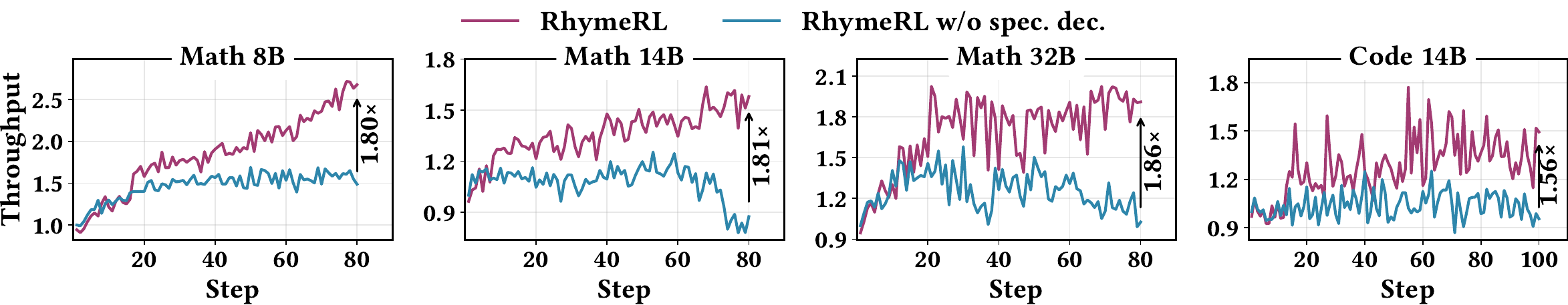}
    \setlength{\belowcaptionskip}{-5pt}
	\setlength{\abovecaptionskip}{-7pt}
    \caption{\textbf{\histospec's rollout throughput improvements.}
    \emph{Max response length\,=\,8K. The data are normalized to the baseline's first step.}
    }
    \label{fig:eval-spec}
\end{figure*}

\subsubsection{Detailed Analysis of \histospec}\
\label{sec:eval-spec}

\myparagraph{Improvement across steps.} 
We present \histospec's rollout throughput gains in each step at 8K max response length in \autoref{fig:eval-spec}.
\histospec delivers up to 1.86x per-step rollout throughput gains, with progressive enhancement observed throughout training. This acceleration stems from: (1) increasing proportion of tokens generated by speculation and acceptance rate, and (2) growing decoding dominance due to extending response lengths, which exacerbates memory-bound limitations. 

\begin{figure}[t]
    \centering
    \includegraphics[width=\linewidth]{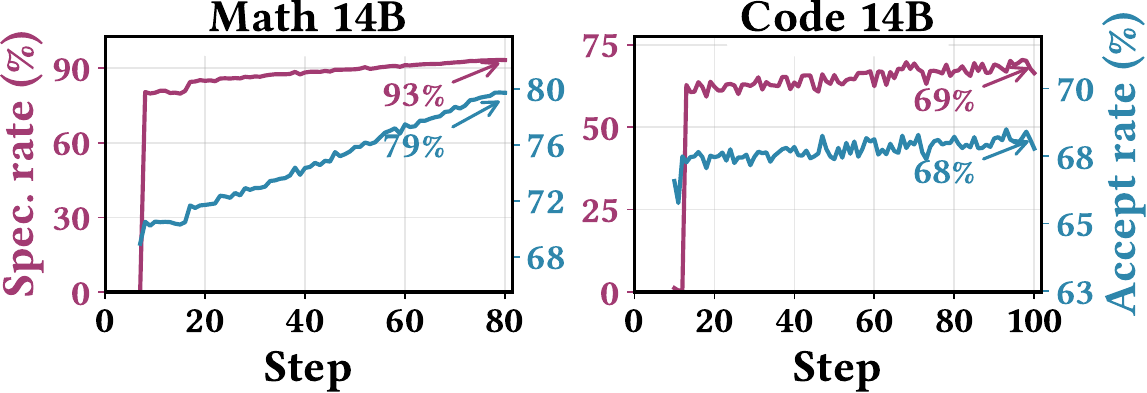}
    \setlength{\belowcaptionskip}{-5pt}
	\setlength{\abovecaptionskip}{-10pt}
    \caption{\textbf{Speculation rate and acceptance rate.}
    \emph{}
    }
    \label{fig:eva-accept-rate}
\end{figure}

\myparagraph{Speculation rate and acceptance rate.}
As shown in \autoref{fig:eva-accept-rate}, the proportion of tokens generated from speculative decoding (speculation rate) increases as training continues.
The acceptance rate, i.e., the proportion of accepted tokens among the draft, remains 65\%--79\% and also increases as training continues.
Utilizing the reward-aware tree-based history management (\autoref{sub:design:spec-tree}) and AIMD-like token speculation (\autoref{sub:design:AIMD}), \histospec achieves high speculation rate and acceptance rate without wasting much computational resources.

\subsubsection{Detailed Analysis of \histopipe}\
\label{sec:eval-pipe}

\myparagraph{Improvement across steps.}
We evaluate the improvements of \histopipe with/without Two-tier Scheduling across steps.
As \autoref{fig:eva-pipe} shows, \histopipe shortens the training time per 10 steps by up to 1.68x, and the Two-tier Scheduling (\autoref{sub:design:adaptive-parallelism}) accelerates the naive hybrid pipe by up to 1.14x. As the training progresses, the response length distribution becomes more stable, and the improvement becomes more significant.

\begin{figure}[t]
    \centering
    \includegraphics[width=\linewidth]{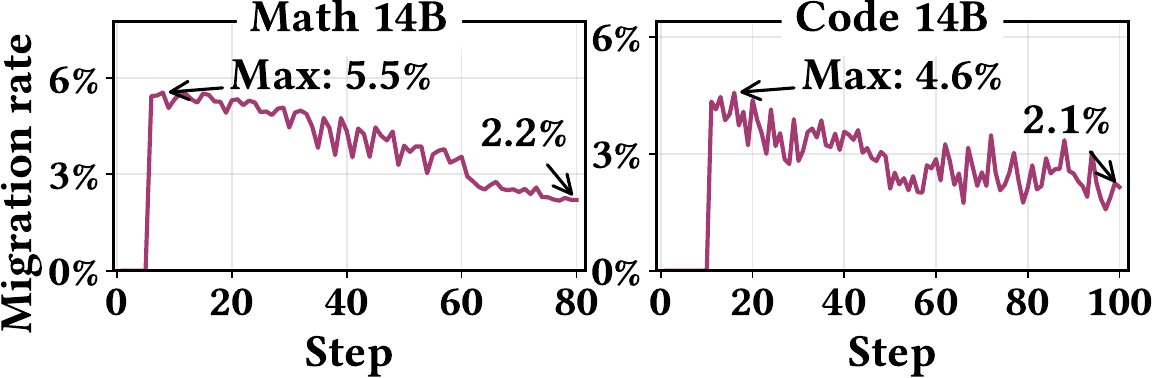}
    \setlength{\belowcaptionskip}{-5pt}
	\setlength{\abovecaptionskip}{-10pt}
    \caption{\textbf{The proportion of samples migrated.}
    \emph{}
    }
    \label{fig:eval-trunc}
\end{figure}

\myparagraph{Migration Rate.}
We quantify the proportion of rollout samples migrated (intra-step or inter-step, \autoref{sub:design:migration-truncation}) as outliers across steps. 
We use 16 ranking groups for math models and 8 ranking groups for code models. As \autoref{fig:eval-trunc} shows, for math tasks, 2.2\%---5.5\% of samples undergo migration (1.6\%---4.6\% for code tasks), with the outlier proportion progressively decreasing during training. 
Such minor migration maintains the efficiency of \histopipe, while preserving algorithmic integrity (\autoref{sec:accuracy}) and introducing ignoreable overhead.

\subsection{Accuracy and Algorithmic Integrity}
\label{sec:accuracy}

\begin{figure}[t]
    \centering
    \includegraphics[width=\linewidth]{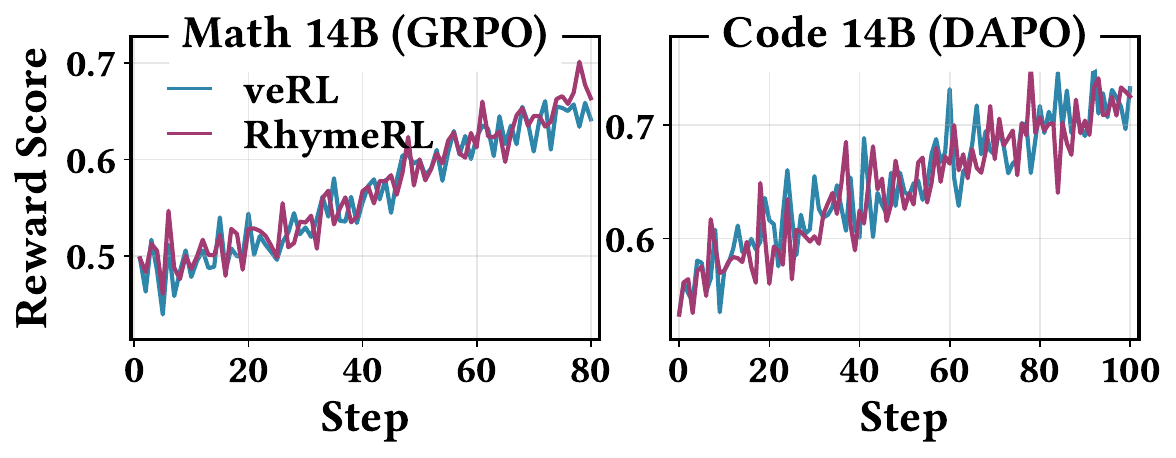}
    \setlength{\belowcaptionskip}{-5pt}
	\setlength{\abovecaptionskip}{-10pt}
    \caption{\textbf{The reward scores of Math and Code 14B LLMs.}
    \emph{}
    }
    \label{fig:eval-accuracy}
\end{figure}

We present the reward scores during RL training.
As \autoref{fig:eval-accuracy} shows, for both GRPO and DAPO training, the curve of \sys closely overlaps with that of veRL.
\sys maintains the overall training accuracy because:
(1) Speculative decoding is theoretically proven to guarantee output equivalence.
(2) The off-policyness of \sys is strictly limited to one step, which is proven to maintain training accuracy~\cite{deepscaler2025, zhong2025streamrlscalableheterogeneouselastic,verl-one-step-off}.
(3) Leveraging historical distribution, \sys only inter-step migrated a small portion of the rollout samples and continues their rollouts in the next step.

\subsection{\histospec vs. Model-based Speculation}
\label{sec:compare-with-other}

Although model-based speculation methods can also accelerate LLM inference, they face significant challenges in LLM RL rollout, making \histospec outperform them. 

\myparagraph{Draft model adaptability.}
Small LLMs cannot have peer reasoning capabilities as large LLMs~\cite{kaplan2020scalinglawsneurallanguage}; 
therefore, SOTA methods utilize the target LLM's hidden states for draft generation~\cite{li2025eaglespeculativesamplingrequires,li2024eagle2fasterinferencelanguage,li2025eagle3scalinginferenceacceleration,cai2024medusasimplellminference}.
However, they still face fundamental challenges in RL due to continuous model evolution.
During RL, the LLMs are updated iteratively for thousands of steps.
However, SOTA methods typically acquire draft models through distillation from the static model (Eagle1-3~\cite{li2025eaglespeculativesamplingrequires, li2024eagle2fasterinferencelanguage,li2025eagle3scalinginferenceacceleration}) or incremental fine-tuning of frozen base models (Medusa~\cite{cai2024medusasimplellminference}), limiting their adaptability to the dynamically evolving LLMs.
While concurrently training the draft model is possible, we have observed that, in practice, due to the effects of \emph{algorithmic stochasticity and uncertainty}, such concurrently trained draft models often fail to deliver high-accuracy drafts consistently throughout the entire training process.

\myparagraph{Performance.} 
\histospec outperforms SOTA model-based methods, stemming from three key advantages: (1) ultra-low draft generation overhead (hundreds of CPU cycles vs. millisecond-level costs in model-based speculation), (2) zero GPU computation or HBM footprint for drafting, and (3) consistently high acceptance rates.
SOTA methods like Eagle3~\cite{li2025eagle3scalinginferenceacceleration} suffer throughput degradation beyond a batch size of 64 despite using a single-layer transformer (also reported by Eagle3~\cite{li2025eagle3scalinginferenceacceleration} and vLLM~\cite{eagle3-worse}).
In contrast, \histospec sustains significant acceleration even at extreme batch sizes.
At step 80, automatic oversampling
leads to batch sizes of 4,928 rollouts/TP for Math-32B and 2,176 rollouts/TP for Math-8B, where \histospec still accelerates rollout by 1.80x---1.86x.

\section{Conclusion}
\label{s:conclusion}
We present \sys, an LLM RL system leveraging historical similarity for efficiency optimization.
As the first RL system leveraging speculative decoding to shorten rollout time, \sys utilizes historical rollout sequences as draft sources.
Leveraging historical distribution, it proposes distribution-aware scheduling to reduce rollout bubbles.
\sys improves RL efficiency without compromising accuracy.

\bibliographystyle{ACM-Reference-Format} %
\bibliography{ref}

\end{document}